%% file: top_and_sup.tex
\documentclass[hidelinks]{article}

\usepackage[nonatbib,final]{neurips_2021}

\usepackage[utf8]{inputenc} %
\usepackage[T1]{fontenc}    %
\usepackage{url}            %
\usepackage{booktabs}       %
\usepackage{nicefrac}       %
\usepackage{microtype}      %

\usepackage{times}
\usepackage{epsfig}
\usepackage{graphicx}
\usepackage{amsmath,amsfonts,amssymb,amsthm}
\usepackage{color}
\usepackage{caption}
\usepackage{subcaption}
\usepackage{wrapfig}
\usepackage{xspace}
\usepackage{array}
\usepackage{cite}
\usepackage{multirow}
\usepackage{booktabs}
\usepackage{pifont} %
\usepackage[thicklines]{cancel}
\usepackage{enumitem}
\usepackage{xcolor}         %

\usepackage[pagebackref=true,breaklinks=true,letterpaper=true,colorlinks,bookmarks=false]{hyperref}

\input{shortcuts}

\begin{document}

\title{On the Frequency Bias of Generative Models}

\author{%
  Katja Schwarz\\
   \And
   Yiyi Liao \\
   \And
   Andreas Geiger \\
   \AND
   \vspace{-2em}{}\\
   Autonomous Vision Group \\
   University of Tübingen and MPI for Intelligent Systems\\
   \texttt{\tt\small {\{firstname.lastname\}}@uni-tuebingen.de} \\
}

\maketitle
\linepenalty=1000

\input{sec_abstract}
\input{sec_intro}
\input{sec_preliminaries}
\input{sec_generator}

\input{sec_discriminator}
\input{sec_gans}
\input{sec_related}

\input{sec_conclusion}
\input{sec_acknowledgments}

{\small
\bibliographystyle{ieee}
\bibliography{bibliography_long,bibliography,bibliography_custom}
}

\newpage
\appendix
\input{supplementary/sec_testbeds.tex}
\input{supplementary/sec_generator.tex}
\input{supplementary/sec_discriminator.tex}
\input{supplementary/sec_gans.tex}

\input{supplementary/sec_datasets.tex}

\end{document}

%% file: shortcuts.tex
\newcommand{\bI}{\mathbf{I}}

\newcommand{\bS}{\mathbf{S}}

\newcommand{\bx}{\mathbf{x}}

\newcommand{\bz}{\mathbf{z}}

\newcommand{\nE}{\mathbb{E}}

\newcommand{\nR}{\mathbb{R}}

\newcommand{\cD}{\mathcal{D}}

\newcommand{\figref}[1]{Fig.~\ref{#1}}
\newcommand{\secref}[1]{Section~\ref{#1}}

\newcommand{\tabref}[1]{Table~\ref{#1}}

\DeclareMathOperator{\atantwo}{atan2}

\makeatletter
\DeclareRobustCommand\onedot{\futurelet\@let@token\@onedot}
\def\@onedot{\ifx\@let@token.\else.\null\fi\xspace}
\def\eg{e.g\onedot} 
\def\ie{i.e\onedot} 
\def\cf{cf\onedot}

\def\wrt{wrt\onedot}

\def\etal{et~al\onedot}

\makeatother

\newcommand{\boldparagraphnospace}[1]{\noindent{\bf #1:} }
\newcommand{\boldparagraph}[1]{\vspace{-0.1cm}\noindent{\bf #1:}\looseness=-1 }
\newcommand{\boldparagraphwocolon}[1]{\vspace{-0.1cm}\noindent{\bf #1}\looseness=-1 }

\definecolor{darkgreen}{rgb}{0,0.7,0}

%% file: sec_abstract.tex
\begin{abstract}
The key objective of Generative Adversarial Networks (GANs) is to generate new data with the same statistics as the provided training data. 
However, multiple recent works show that state-of-the-art architectures yet struggle to achieve this goal. 
In particular, they report an elevated amount of high frequencies in the spectral statistics which makes it straightforward to distinguish real and generated images.
Explanations for this phenomenon are controversial: While most works attribute the artifacts to the generator, other works point to the discriminator.  
We take a sober look at those explanations and provide insights on what makes proposed measures against high-frequency artifacts effective.
To achieve this, we first independently assess the architectures of both the generator and discriminator and investigate if they exhibit a frequency bias that makes learning the distribution of high-frequency content particularly problematic.
Based on these experiments, we make the following four observations: 
1) Different upsampling operations bias the generator towards different spectral properties.
2) Checkerboard artifacts introduced by upsampling cannot explain the spectral discrepancies alone as the generator is able to compensate for these artifacts.
3) The discriminator does not struggle with detecting high frequencies per se but rather struggles with frequencies of low magnitude. 
4) The downsampling operations in the discriminator can impair the quality of the training signal it provides.
In light of these findings, we analyze proposed measures against high-frequency artifacts in state-of-the-art GAN training but find that none of the existing approaches can fully resolve spectral artifacts yet.
Our results suggest that there is great potential in improving the discriminator and that this could be key to match the distribution of the training data more closely.
\end{abstract}

%% file: sec_intro.tex
\section{Introduction}
\noindent
In recent years, unconditional Generative Adversarial Networks (GANs) have achieved impressive photo-realism for image synthesis tasks. 
While this has hampered the identification of generated images based on visual cues, multiple recent works show that it is straightforward to distinguish real and generated images based on their high-frequency content~\cite{Jiang2020ARXIV,Wang2020CVPR,Dzanic2020NIPS,Durall2020CVPR,LiuZ2020CVPR,Zhang2019IEEE,Frank2020ICML}. This has aroused considerable interest because it reveals a fundamental problem in state-of-the-art GANs:
Existing approaches evidently struggle to learn the correct data distribution. 
While GAN training is notoriously hard, learning the distribution of high-frequency content is particularly problematic~\cite{Jiang2020ARXIV,Wang2020CVPR,Dzanic2020NIPS,Durall2020CVPR,LiuZ2020CVPR,Zhang2019IEEE,Frank2020ICML}.
This indicates a systematic problem in existing approaches that could make training suboptimal. 
For example, if generating high-frequencies is difficult for the generator but detecting them is straightforward for the discriminator this imbalance could impair the stability of the training.
Conversely, if the discriminator struggles to detect high frequencies generating fine details also becomes more difficult which could impede convergence.
Therefore, we argue that it is important to better understand the expressivity of both the generator and the discriminator.
Indeed, numerous works suggest that the architecture of the generator can hamper the generation of high frequencies~\cite{Chandrasegaran2021CVPR,Durall2020CVPR,Frank2020ICML,Khayatkhoei2020ARXIV,Jiang2020ARXIV,Gal2021ARXIV}.
However, another line of works question the quality of the training signal provided by the discriminator~\cite{Jung2021AAAI,Chen2021AAAI,Gal2021ARXIV}.
But is there indeed a \textit{frequency bias} that prevents learning high frequencies in existing GAN models? \\
This is a non-trivial question as GAN training involves two players where architectures for both the generator and discriminator, loss functions, as well as the dataset statistics can all affect the generated images. To narrow down potential factors, we first develop isolated testbeds for both the generator and discriminator.
Then, we extend our findings to state-of-the-art GANs on large-scale datasets.
\begin{figure}[t!]
  \centering
  \begin{subfigure}[t]{0.16\linewidth}
  \includegraphics[height=\linewidth]{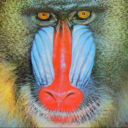}\\
  \includegraphics[height=\linewidth]{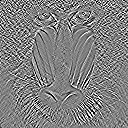}
  \caption{Ground truth}
  \end{subfigure}
  \hfill
  \begin{subfigure}[t]{0.4\linewidth}
  \includegraphics[height=0.4\linewidth]{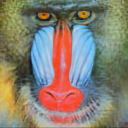}
  \includegraphics[height=0.4\linewidth]{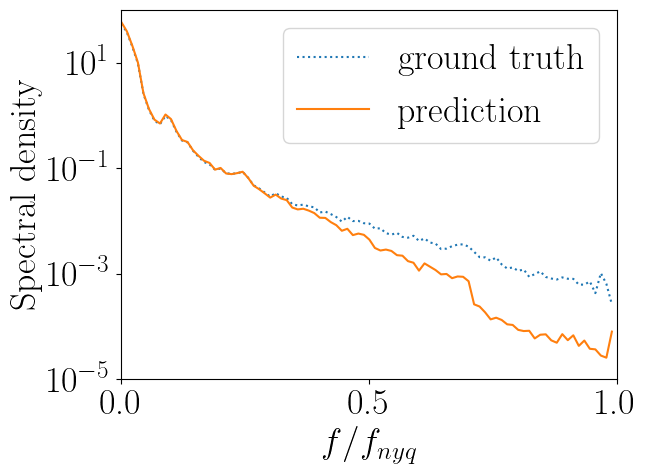}\\
   \includegraphics[height=0.4\linewidth]{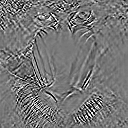}
  \includegraphics[height=0.4\linewidth]{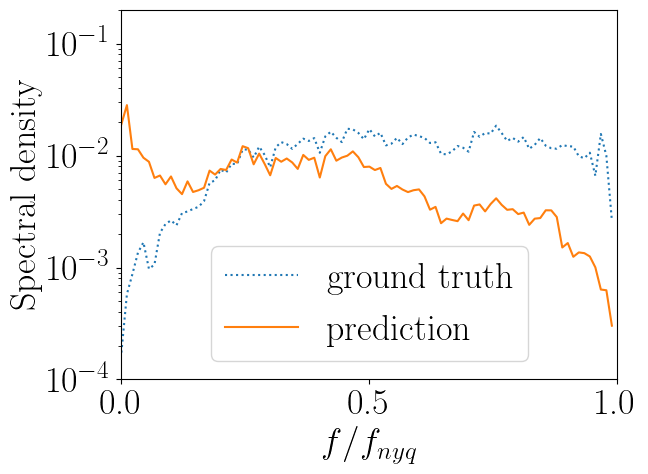}
   \caption{Reconstruction of a generator with bilinear upsampling using pixel-level supervision}
   \label{fig:teaser_generator}
   \end{subfigure}
    \hfill
  \begin{subfigure}[t]{0.4\linewidth}
    \includegraphics[height=0.4\linewidth]{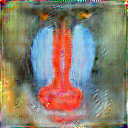}
	\includegraphics[height=0.4\linewidth]{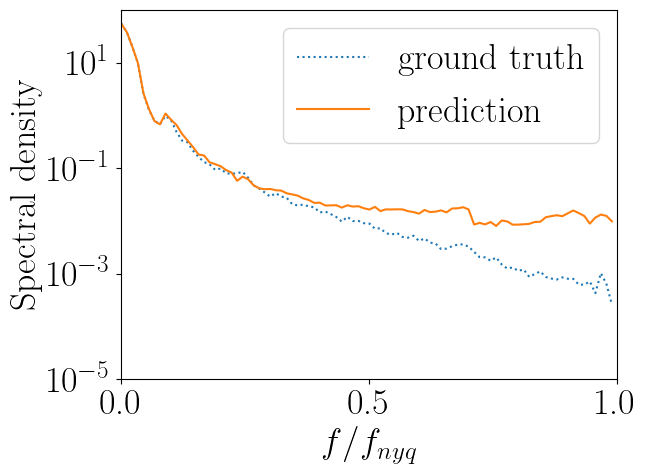}\\
	\includegraphics[height=0.4\linewidth]{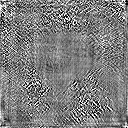}
	\includegraphics[height=0.4\linewidth]{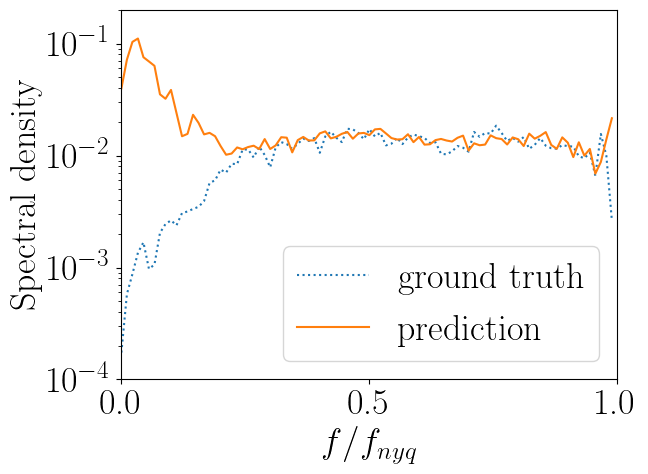}
  \caption{Reconstruction guided by a discriminator with BlurPool downsampling}  
   \label{fig:teaser_discriminator}
   \end{subfigure}
  \caption{\textbf{Spectral Properties of Generator and Discriminator.} (a) We show a natural image and a high-pass filtered version (amplified for clarity) to illustrate the networks' behavior for both low and high magnitudes at high frequencies. 
  (b) Our experiments reveal that bilinear upsampling biases the generator towards generating data with little high-frequency content, regardless of the magnitude.
  (c) The discriminator shows no such bias but provides better guidance for frequencies with high magnitude. The results further reveal artifacts due to downsampling that impair the training signal.
  We include the training details for both experiments in the supplementary.}
  \label{fig:teaser}
  \vspace{-0.5cm}
\end{figure}

\boldparagraphnospace{Generator} 
The majority of works attribute the high-frequency artifacts to the upsampling operations in the generator~\cite{Chandrasegaran2021CVPR,Durall2020CVPR,Frank2020ICML,Odena2016Distill,Zhang2019IEEE}. 
Upsampling typically follows a pre-defined scheme, e.g. bilinear or nearest neighbor interpolation, or zero insertion between pixels. 
While the latter introduces too many high frequencies, interpolation has been shown to reduce artifacts~\cite{Chandrasegaran2021CVPR,Durall2020CVPR}. 
But \textit{is interpolation always a good scheme for upsampling?} 
Our experiments indicate that both bilinear and nearest neighbor upsampling bias the generator towards predicting little high-frequency content, see \figref{fig:teaser_generator}.
While zero insertion can be more flexible, it is prone to introducing checkerboard artifacts, \ie too many high frequencies~\cite{Durall2020CVPR,Chandrasegaran2021CVPR,Odena2016Distill}. But \textit{why do the learnable filters in subsequent layers not learn to compensate for these artifacts?}
Depending on the training objective, checkerboard artifacts might be penalized only slightly. Our experiments evidence that when the loss function is sensitive to these artifacts the generator is indeed capable of compensating for them.

\boldparagraphnospace{Discriminator} 
This leads us to question the quality of the training signal: \textit{Can the discriminator detect high frequencies and provide the necessary supervision?} 
Chen \etal~\cite{Chen2021AAAI} argue that the discriminator cannot detect high-frequency information due to the downsampling operations.
Our experiments corroborate that downsampling can introduce artifacts in the training signal. However, we also observe that the discriminator can indeed provide a meaningful training signal for the spectral statistics at high frequencies when their magnitude is large enough, \eg, \figref{fig:teaser_discriminator}. 
Other works propose to train both the generator and discriminator in wavelet space~\cite{Gal2021ARXIV} or add a discriminator on the spectrum of the images to reduce high-frequency artifacts~\cite{Jung2021AAAI}.
Motivated by these approaches we ask: \textit{Is it enough to consider only the spatial domain?}
In agreement with these works, our results suggest that the discriminator can benefit from (additional) input in frequency-based domains. 
Further, our testbed yields insights on which of these measures is most effective and reveals that the training signal from the discriminator might remain problematic.

\boldparagraphnospace{Contributions} 
We take a sober look at explanations for high-frequency artifacts in generated images and unify the efforts that have been done so far. In particular, we develop isolated testbeds for both the generator and the discriminator which allows us to analyze what makes existing efforts effective.
The conclusions drawn by this paper shed new light on limitations of common design choices for both generator and discriminator: 
i) Bilinear and nearest neighbor upsampling bias the generator towards predicting little high-frequency content.
ii) Zero insertion is prone to producing checkerboard artifacts in the generated images. However, with a suitable loss function, the learnable filters of the generator can compensate for the artifacts. This indicates that the upsampling in the generator alone cannot explain the spectral discrepancies.
iii) In general, the discriminator is able to detect high frequencies and provide supervision to learn the correct spectral statistics. However, while the exponential decay of the spectrum for natural images creates the impression of the discriminator being insensitive to high frequencies, it actually struggles with low magnitudes.
iv) We find that all commonly used downsampling operations in the discriminator can impair the quality of the training signal.

Lastly, we demonstrate that these findings extend to full GAN training.  
In agreement with~\cite{Jung2021AAAI}, we find that spectral discriminators can bring us one step closer to matching the spectral statistics. Nonetheless, generated images remain straightforward to classify based on their spectral statistics only. 
While recent works on GANs largely focus on improving the generator, \eg,~\cite{Karras2018ICLR,Karras2019CVPR,Karras2020CVPR}, our findings suggest that the design of the discriminator plays an equally important role and deserves more attention in future work. 
We believe that our testbeds for both the generator and discriminator can be a useful tool for future investigations. We release our code and dataset at \url{https://github.com/autonomousvision/frequency_bias}.

%% file: sec_preliminaries.tex
\section{Preliminaries}\label{sec:preliminaries}
\boldparagraph{GAN Training}
A generative adversarial network consists of a generator and a discriminator that are trained jointly in a 2-player game. 
Given latent variables $\bz\sim p_z$, the generator synthesizes images while the discriminator tries to distinguish the synthesized images from real images $\bI\sim p_\cD$, sampled from data distribution $p_\cD$.
Let $G_\Theta$ and $D_\Phi$ denote a generator $G$ and a discriminator $D$ with parameters $\Theta$ and $\Phi$, respectively. The parameters of the models are updated in alternating steps using a non-saturating GAN objective~\cite{Goodfellow2014NIPS} which is often combined with R1-regularization to stabilize training \cite{Mescheder2018ICML}
\begin{equation}
V(\Theta, \Phi) =
\nE_{\bz\sim p_{z}}
\left[f(D_\Phi(G_\Theta(\bz)))\right]
\,+\, \nE_{\bI \sim p_{\cD}}
\left[
f(-D_\Phi(\bI))
\,-\, \lambda {\Vert \nabla D_\Phi(\bI)\Vert}^2
\right]
\end{equation}
where $f(t)=-\log(1+\exp(-t))$ and $\lambda$ controls the strength of the regularizer. 

\boldparagraph{Image Processing in the Frequency Domain}
The discrete 2D Fourier transform maps a gray scale image $\bI\in\nR^{H \times W}$ to the frequency domain:
\begin{equation}
	\hat{\bI}[k,l] = \frac{1}{HW} \sum_{x=0}^{H-1}\sum_{y=0}^{W-1} \exp^{-2\pi i \frac{x\cdot k}{H}}\exp^{-2\pi i \frac{y\cdot l}{W}} \cdot \bI[x, y]
\end{equation}
for $k=0,\dots,H-1$ and $l=0,\dots,W-1$. 
The power spectral density is estimated by the squared magnitudes of the Fourier components $\bS[k,l] = |\hat{\bI}[k,l]|^2$. 
Similar to~\cite{Durall2020CVPR,Dzanic2020NIPS} we consider the reduced spectrum~$\tilde{S}$, i.e. the azimuthal average over the spectrum in normalized polar coordinates $r\in[0,1]$, $\theta\in[0,2\pi)$
\begin{equation}
	\tilde{S}(r) = \frac{1}{2\pi}\int_0^{2\pi} S(r,\theta) d\theta \quad \text{with} \quad
	r = \sqrt{\frac{k^2+l^2}{\frac{1}{4}(H^2 + W^2)}}\quad \text{and} \quad \theta=\atantwo(k,l) \\
\label{eq:freqradius}
\end{equation}
Since images are discretized in space, the maximum frequency is determined by the Nyquist frequency. %
For a square image, $H=W$, it is given by $f_{nyq}=\sqrt{k^2+l^2}=H/\sqrt{2}$, \ie for $r=1$.

\boldparagraph{Spectral Classifier}
To detect generated images, Dzanic \etal~\cite{Dzanic2020NIPS} propose to classify real and generated images based on their reduced spectrum. More specifically, they fit a power-law function to the tail of each reduced spectrum for frequencies above a given threshold $r_c=0.75$ and train a binary classifier on the fit parameters of each spectrum.

%% file: sec_generator.tex
\section{Are high frequencies more difficult to generate?}\label{sec:generator}
In this section, we investigate if there is a frequency bias in the generator that impedes the generation of high frequencies. Therefore, we first consider the generator in an isolated setting assuming pixel-level supervision -- otherwise, even with a perfect discriminator it would be impossible to generate a correct image. We extend our findings to the full GAN setting in ~\secref{sec:gans}.\\ 
There are two lines of argumentation in existing works: The first investigates artifacts that arise from the upsampling operations in the generator~\cite{Chandrasegaran2021CVPR,Durall2020CVPR,Frank2020ICML}. These works analyze the spectral properties of the generator predictions at convergence. We extend this analysis from the perspective of a frequency bias to see if low frequencies are learned earlier during training. 
The second line of works argues for a frequency bias of the learnable filters~\cite{Khayatkhoei2020ARXIV,Jiang2020ARXIV}. Motivated by their analysis we investigate if learnable filters are at all able to compensate for the artifacts introduced by upsampling.

\boldparagraph{Experimental Setting}
To isolate the generator from the discriminator we consider a conditional reconstruction task. In particular, we take 10 images from a dataset and pair them with 10 latent codes drawn from a normal distribution. Given a latent code, the generator is optimized to reconstruct the corresponding image with a pixel-wise L2-loss
\begin{equation}
L_I = \frac{1}{HW} \sum_{x=0}^{H-1}\sum_{y=0}^{W-1} \left \| G_\Theta(\bz)[x,y] - \bI[x,y] \right \|_2^2
\end{equation}
We choose the generator from PGAN~\cite{Karras2018ICLR} because of its simple architecture comprising only convolutions and upsampling operations. Further, when combined with R1-regularization it trains stably in a GAN setting~\cite{Mescheder2018ICML}, allowing us to use the same generator in~\secref{sec:gans}. We reduce the number of channels for faster training because our primary focus is on the spectral properties. As upsampling operations, we investigate bilinear and nearest neighbor interpolation, zero insertion, and reshaping in the channel dimension~\cite{ShiCVPR2016}. We ensure that all networks have a similar number of parameters. More details are provided in the supplementary.

\begin{figure}[t!]
  \centering
	 \begin{subfigure}[b]{0.16\linewidth}
   	 	{\raisebox{0.8cm}{\rotatebox[origin=l]{90}{\small{Toyset}}}} 
	  	\raisebox{0.5cm}{
	 	\includegraphics[width=0.8\linewidth]{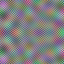}}\\
	 	{\raisebox{0.2cm}{\rotatebox[origin=l]{90}{\small{CelebA}}}}
	 	\raisebox{0cm}{
	 	\includegraphics[width=0.8\linewidth]{gfx/generator/celeba_64/train_sample}}
	 \vspace{0.5cm}
	 \caption{GT Sample}
	 \label{fig:generator_gt}
	 \end{subfigure}
	  \begin{subfigure}[b]{0.223\linewidth}
	 	\includegraphics[width=\linewidth]{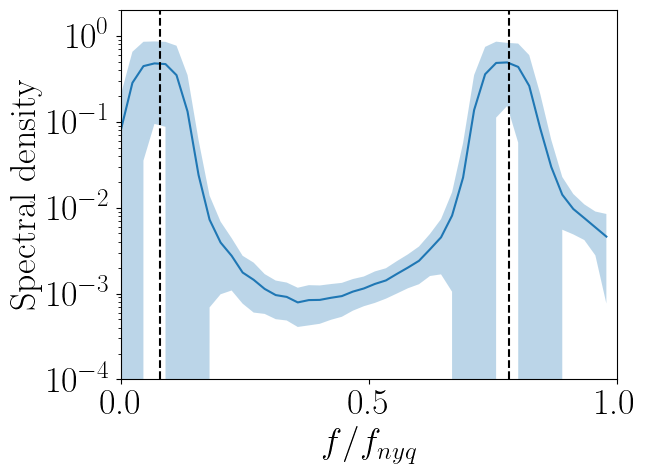}\\
	 	\includegraphics[width=\linewidth]{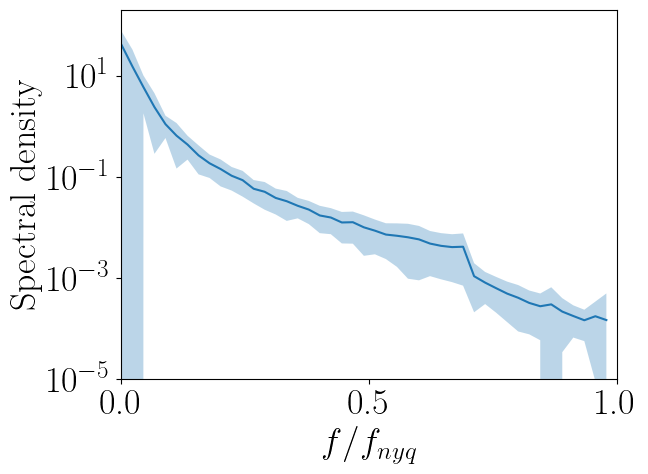}
	 \caption{GT Spectrum}
	\label{fig:generator_spectrum}
	 \end{subfigure}
	 \begin{subfigure}[b]{0.16\linewidth}
	 	\includegraphics[width=\linewidth]{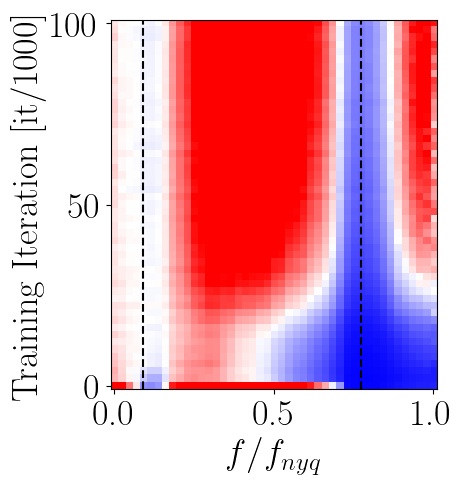}\\
	 \includegraphics[width=\linewidth]{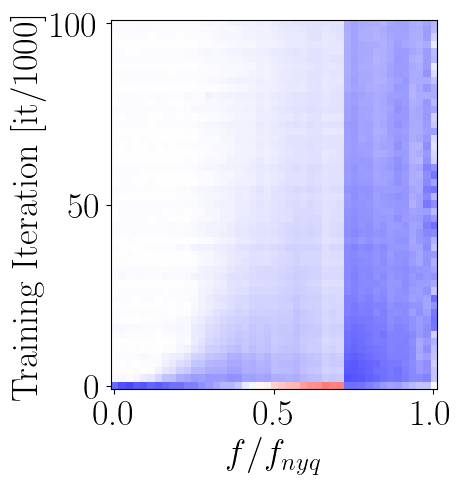}
	 	 \caption{Bilinear}
	 	 	 \label{fig:generator_bilinear}
	 \end{subfigure}
	 \begin{subfigure}[b]{0.132\linewidth}
	 	\includegraphics[width=\linewidth]{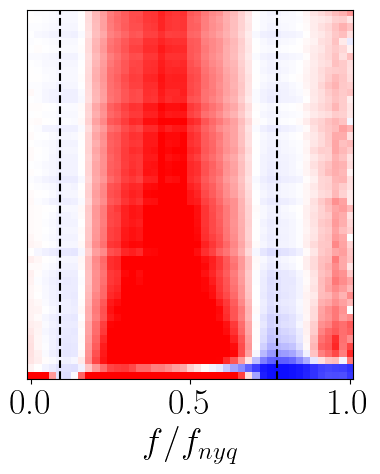}
		 \includegraphics[width=\linewidth]{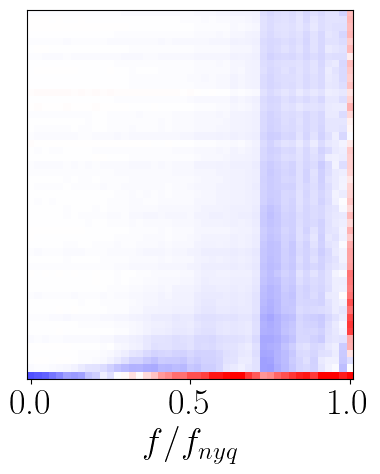}
	 	 \caption{NN}
	 	 	 \label{fig:generator_nn}
	 \end{subfigure}
	 \begin{subfigure}[b]{0.132\linewidth}
	 	\includegraphics[width=\linewidth]{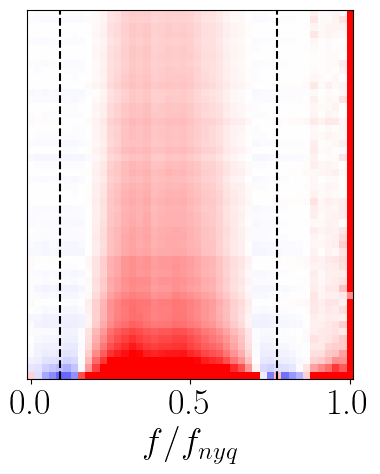}\\
	 	\includegraphics[width=\linewidth]{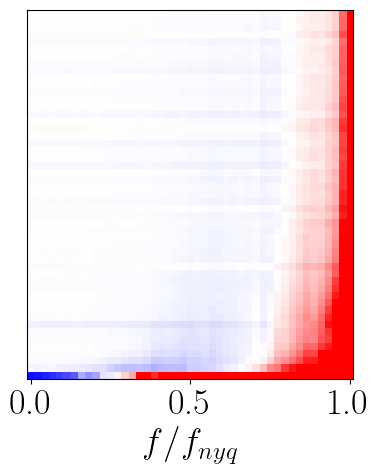}
	\caption{Zeros}
		 \label{fig:generator_zeros}
	\end{subfigure}
	\begin{subfigure}[b]{0.151\linewidth}
		\includegraphics[width=\linewidth]{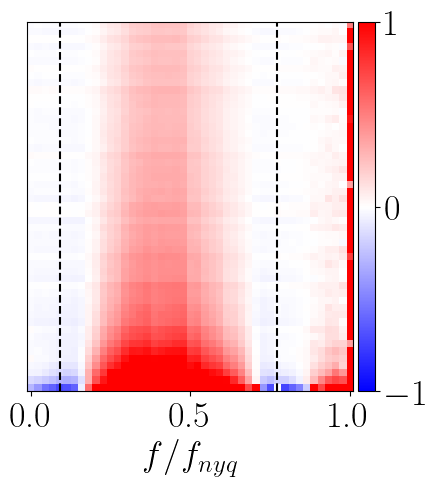}\\
	 	\includegraphics[width=\linewidth]{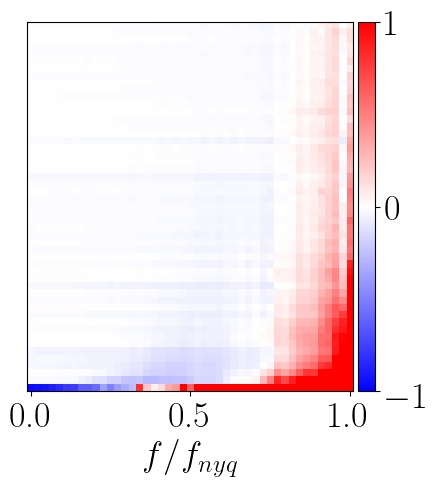}
	 \caption{Reshape}
	 	 \label{fig:generator_reshape}
	 \end{subfigure}
    \caption{\textbf{Spectrum Error Evolution for the Generator.} We show one sample from the dataset in the first column and the mean and standard deviation of the reduced spectra for all 10 samples in the second column for reference. For the Toyset the dashed lines mark the mean of the Gaussian peaks at $0.1f_{nyq}$ and $0.8f_{nyq}$.
    Upsampling with bilinear or nearest neighbor interpolation biases the generator towards predicting little high-frequency content. Conversely, zero insertion and reshaping are prone to introducing checkerboard artifacts, indicated by the large errors at the highest frequency. The color corresponds to the relative error of the average predicted reduced spectrum \wrt the ground truth and is clipped at $1$, i.e. when the relative error exceeds $100\%$.}
  \label{fig:generator}
  \vspace{-0.3cm}
\end{figure}

\boldparagraph{Datasets}
In natural images, the spectral density follows an exponential decay. To isolate the effects of frequency range and magnitute, we create a Toyset of images with two Gaussian peaks $\mathcal{N}(\mu_1,\sigma_1)$, $\mathcal{N}(\mu_2,\sigma_2)$ of equal magnitude in the spectrum, see \figref{fig:generator_gt},~\ref{fig:generator_spectrum}.
Based on the Nyquist frequency $f_{nyq}=H/\sqrt{2}$, we create samples using $\sigma_1=\sigma_2=1/\sqrt{2}f_{nyq}$ and draw $\mu_1\sim \mathcal{U}(0.05f_{nyq},0.15f_{nyq})$ and $\mu_2\sim \mathcal{U}(0.75f_{nyq}, 0.85f_{nyq})$. Together with a uniformly distributed phase, we apply the inverse Fourier transform to create images.
We further test our setting on natural images with a downsampled version of CelebA~\cite{Liu2015ICCVa}. Both datasets have a resolution of $64^2$ pixels.

\boldparagraph{Evaluation Metrics}
On the spatial domain, we report PSNR on the RGB values and on the frequency domain we evaluate the reduced spectrum, see Eq.~\eqref{eq:freqradius}. To analyze frequency-dependent convergence we further visualize the evolution of the spectrum similar to \cite{Rahaman2019ICML}. Here, the x-axis denotes the (frequency) radius $r$ in normalized polar coordinates and the y-axis denotes the training iterations.
The color corresponds to the relative error of the average predicted reduced spectrum \wrt the ground truth, where positive and negative values indicate too many and too few predicted frequencies, respectively. We clip the colorbar at $1$, \ie, when the relative error exceeds $100\%$.

\boldparagraphwocolon{Do generators exhibit a frequency bias?}
The spectral evolution in \figref{fig:generator} shows different behavior for bilinear and nearest neighbor upsampling compared to zero insertion and reshaping. 
Particularly on the Toyset, a generator with bilinear or nearest neighbor upsampling learns the lower frequencies earlier in training than the high frequencies, see~\figref{fig:generator_bilinear} and~\ref{fig:generator_nn}. 
While the generator with bilinear upsampling struggles with high frequencies throughout training, for nearest neighbor upsampling it eventually fits the high-frequency peak, suggesting that its bias towards little high-frequency content is not as strong.
In contrast, a generator with zero insertion or reshaping learns both peaks approximately at equal speed but is prone to generating checkerboard artifacts as indicated by the large error at the highest frequency in~\figref{fig:generator_zeros} and~\ref{fig:generator_reshape}.  
Between the peaks of the toyset, where frequencies have small magnitudes, all methods struggle to match the statistics of the training data because the L2-loss penalizes errors at frequencies with low magnitude only slightly.
This explains why for CelebA there is an overall trend towards learning lower frequencies earlier during training.
The PSNR values in \tabref{tab:generator} support these findings in the spatial domain. The lack of high frequencies for bilinear and nearest neighbor upsampling manifests in a low PSNR, particularly for the Toyset which contains many high frequencies by construction.
\begin{table}
\centering
	\begin{minipage}{0.47\textwidth}
	\centering
  		\setlength\tabcolsep{0.2em}
		\renewcommand{\arraystretch}{1.0}
  		\begin{small}
  		\scalebox{0.9}{
  		\input{tab/generator/psnr.tex}}
  		\end{small}
  		\vspace{0.1cm}
   		\caption{\textbf{PSNR} for different upsampling operations in the generator at resolution $64^2$ pixels.}
    	\label{tab:generator}
    \end{minipage}\hfill
		\begin{minipage}{0.49\textwidth}
	\centering
  		\setlength\tabcolsep{0.2em}
		\renewcommand{\arraystretch}{1.0}
  		\begin{small}
  		\scalebox{0.9}{
  		\input{tab/discriminator/psnr.tex}}
  		\end{small}
  		\vspace{0.1cm}
   		\caption{\textbf{PSNR} for different downsampling operations in the discriminator at resolution $64^2$ pixels.}
    	\label{tab:discriminator}
    \end{minipage}\hfill
  \vspace{-0.3cm}
\end{table}
\begin{figure}[t!]
    \centering
    \begin{subfigure}[b]{0.222\linewidth}
    	\includegraphics[width=\linewidth]{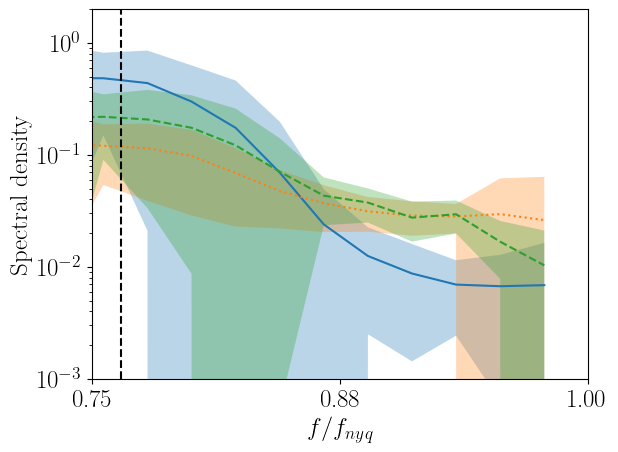}
    	\caption{Bilinear}
        \label{fig:generator_specloss_bilinear}
	\end{subfigure}
    \begin{subfigure}[b]{0.2\linewidth}
    	\includegraphics[width=\linewidth]{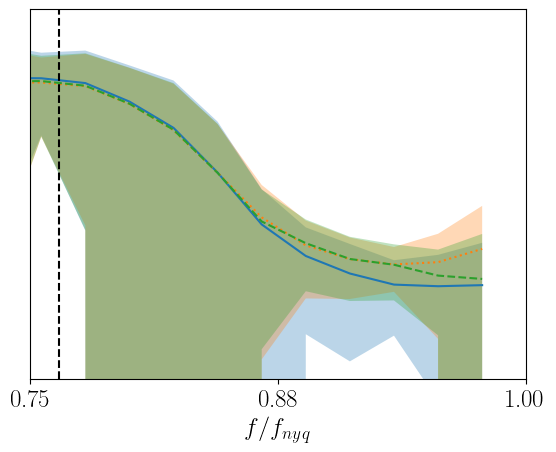}
    	\caption{NN}
     	\label{fig:generator_specloss_nn}
   \end{subfigure}
    \begin{subfigure}[b]{0.2\linewidth}
    	\includegraphics[width=\linewidth]{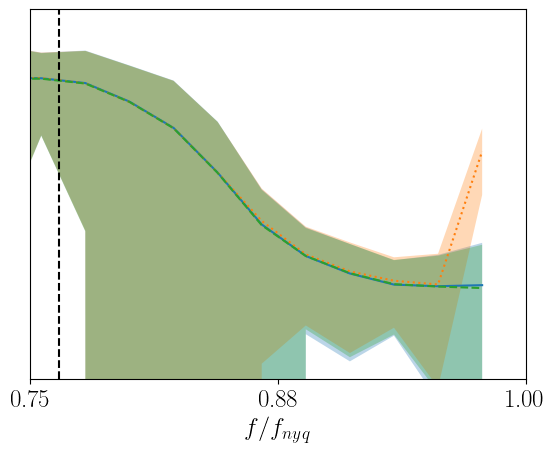}
    	\caption{Zeros}
    	\label{fig:generator_specloss_zeros}
    \end{subfigure}
    \begin{subfigure}[b]{0.2\linewidth}
    	\includegraphics[width=\linewidth]{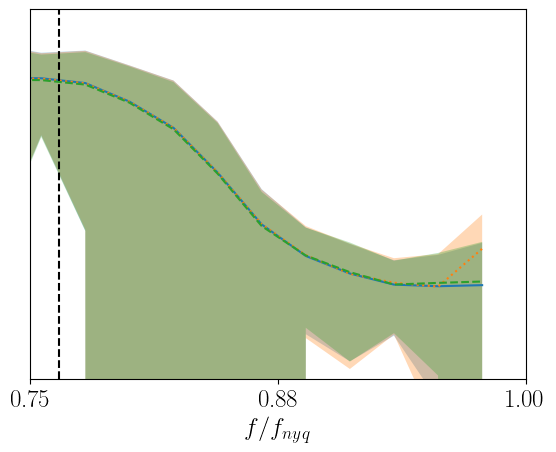}
		\caption{Reshape}
    	\label{fig:generator_specloss_reshape}
    \end{subfigure}
   	\begin{subfigure}[b]{0.14\linewidth}
   	\raisebox{1.3cm}{
    	\includegraphics[width=\linewidth]{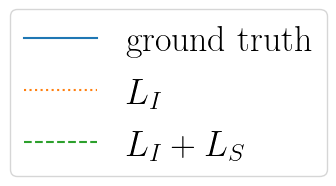}}
    \end{subfigure}
    \caption{\textbf{Reduced Spectrum for the Generator} on the Toyset. 
    We plot the mean and standard deviation of the reduced spectrum above $0.75f_{nyq}$. 
    The peak at the highest frequencies from upsampling with zero insertion or reshaping is removed by an additional loss on the spectrum.}
  \label{fig:generator_specloss}
   \vspace{-0.3cm}
\end{figure}

\boldparagraphwocolon{Can the generator learn to compensate for the artifacts?}
Since the L2-loss is less sensitive to frequencies with low magnitudes (see supplementary for a formal derivation),  we now add an L2-loss on the logarithm of the reduced spectrum:
\begin{equation}
L_S = \frac{1}{H/\sqrt{2}} \sum_{k=0}^{H/\sqrt{2}-1} \left\|\log\left(\tilde{S}\left(G_\Theta(\bz)\right)\right)[k] -\log\left( \tilde{S}(\bI)\right)[k]\right\|_2^2 
\end{equation}
The logarithm penalizes errors at low magnitudes more strongly and can therefore reduce the checkerboard artifacts introduced by zero insertion and reshaping, see \figref{fig:generator_specloss}.
Hence, given a suitable objective, learnable filters can indeed compensate for high-frequency artifacts introduced by zero insertion and reshaping.
Interestingly, bilinear upsampling does not benefit as much from the spectral loss. This suggests that a strong bias towards little high-frequency content can be more difficult to compensate for.
As expected, the additional loss does not alter the evaluation in the spatial domain significantly and yields similar PSNR values in~\tabref{tab:generator}.

\boldparagraph{Implications}
Overall, these results lead us to conclude that different upsampling operations bias the generator towards different spectral properties. Nearest neighbor and particularly bilinear upsampling introduce a bias towards fitting functions with little high-frequency content. 
In \secref{sec:gans}, we will see that this can also be beneficial when working with natural images as their spectral density follows an exponential decay. 
On the other hand, zero insertion and reshaping introduce a bias towards checkerboard artifacts. However, with a suitable loss function, the network filters can learn to compensate for these artifacts. Therefore, the upsampling in the generator alone cannot explain the spectral discrepancies. This suggests that the training signal provided by the discriminator might be suboptimal in the first place.

%% file: tab/generator/psnr.tex
\begin{tabular}{lcccccccc}
\toprule
 & \multicolumn{2}{c}{Bilinear} & \multicolumn{2}{c}{NN} & \multicolumn{2}{c}{Zeros} & \multicolumn{2}{c}{Reshape}\\
 & & + $L_S$ & & + $L_S$ & & + $L_S$ & & + $L_S$\\
\midrule
Toyset &  	21.1 &  	20.9 & 	30.7 & 	30.7 & 	38.6 & 	38.0 & 	38.6 & 36.5 \\
CelebA & 	34.9 & 	37.0 & 	40.9 & 	39.8	 & 	42.4 & 	43.6 & 	42.3 & 40.9 \\

\bottomrule     
\end{tabular}

%% file: tab/discriminator/psnr.tex
\begin{tabular}{lcccccccc}
\toprule
 & \multicolumn{2}{c}{AvgPool} & \multicolumn{2}{c}{BlurPool} & \multicolumn{2}{c}{Stride} & \multicolumn{2}{c}{MLP} \\
 & & + SD & & + SD & & + SD & & +SD\\
\midrule
Toyset &  	16.3 &  	19.2 & 	23.3 & 	15.6 & 	23.3 &  24.1 &	26.2 &	46.5 \\
CelebA & 	25.4 & 	27.0 & 	24.2 & 	24.4 & 	25.5 & 	25.6  &	28.1 &	33.7

 \\
\bottomrule     
\end{tabular}

%% file: sec_discriminator.tex
\section{Can the discriminator provide a good training signal?}\label{sec:discriminator}
In this section, we investigate how good the training signal is that the discriminator can provide. Only a few existing works consider the discriminator as a cause for the spectral discrepancies. Chen \etal~\cite{Chen2021AAAI} attribute the high-frequency artifacts to information loss in the downsampling operations. 
In particular, they analyze how high frequencies affect the output of the discriminator at convergence. Instead, we consider how downsampling affects the training signal by assessing the input to the discriminator. Further, we analyze how the spectral statistics evolve during training to see if downsampling introduces a bias towards correcting low frequencies earlier than high frequencies.
To reduce spectral discrepancies, Chen \etal~\cite{Chen2021AAAI} propose a regularizer based on the reduced spectrum to perform hard example mining in the frequency domain, \cf Eq.~\eqref{eq:freqradius}. Similarly, Jung \etal~\cite{Jung2021AAAI} define an additional discriminator on the reduced spectrum.
While spectral discrepancies are not the main focus of their work, Gal \etal~\cite{Gal2021ARXIV} also argue for "unfavorable loss functions" and propose to train both the generator and discriminator in wavelet space.
In the second part of this section, we therefore aim to understand what makes (additional) inputs from other domains valuable and which of the proposed measures are most effective to correct the high-frequency artifacts.

\boldparagraph{Experimental Setting}
Similar to the generator experiments in~\secref{sec:generator}, we propose a conditional reconstruction task to assess the quality of the training signal independently of the generator architecture. More specifically, we train a class-conditional GAN with a single sample per class. 
To minimize the impact of the generator, we directly optimize the pixel values of the fake images. In practice, we pair 10 images with 10 labels as training data and optimize 10 learnable tensors conditioned on the labels.
We optimize the learnable tensors and discriminator weights in an alternating fashion using the GAN two-player game setting.
Similar to~\secref{sec:generator}, we use the PGAN discriminator~\cite{Karras2018ICLR} with a reduced number of channels because our primary focus is on the spectral properties.
As downsampling operations, we investigate strided convolution, average pooling, and blurring with subsequent average pooling~\cite{Zhang2019ICML}. We further train an MLP on the flattened input image as a baseline without any downsampling operations.
When adding a spectrum discriminator, we weigh both discriminator losses equally as in ~\cite{Jung2021AAAI}.
Note, that training a GAN on such few images is a non-trivial task that requires careful tuning to train stably. We ensured that this is the case for our results, see supplementary for details. 

\boldparagraph{Datasets and Metrics}
We use the same datasets and metrics as in~\secref{sec:generator} and evaluate them on the reconstructions guided by the gradients from the discriminator. This allows us to assess the quality of the training signal.

\begin{figure}[t!]
  \centering
  \begin{subfigure}[t]{0.115\linewidth}
  \includegraphics[width=\linewidth]{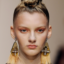}
  \caption{GT}
  \end{subfigure}
  \begin{subfigure}[t]{0.115\linewidth}
  \includegraphics[width=\linewidth]{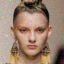}
  \caption{MLP}  
  \end{subfigure}
  \begin{subfigure}[t]{0.115\linewidth}
  \includegraphics[width=\linewidth]{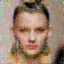}
  \caption{AvgPool}
  \end{subfigure}
  \begin{subfigure}[t]{0.115\linewidth}
  \includegraphics[width=\linewidth]{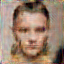}
   \caption{BlurPool}
   \end{subfigure}
  \begin{subfigure}[t]{0.115\linewidth}
  \includegraphics[width=\linewidth]{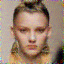}
  \caption{Stride}  
  \label{fig:downsampling_artifacts_stride}
  \end{subfigure}
  \begin{subfigure}[t]{0.115\linewidth}
  \includegraphics[width=\linewidth]{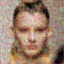}
  \caption{Stride \\+ SD}
  \label{fig:downsampling_artifacts_sd}  
  \end{subfigure}
  \begin{subfigure}[t]{0.115\linewidth}
  \includegraphics[width=\linewidth]{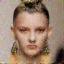}
  \caption{Stride \\+ F-mining}  
  \end{subfigure}
  \begin{subfigure}[t]{0.115\linewidth}
  \includegraphics[width=\linewidth]{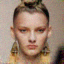}
  \caption{Stride \\+ Wavelet}  
  \end{subfigure}
  \caption{\textbf{Reconstruction Guided by the Discriminator.} 
	All discriminators with downsampling lead to significant deviations from the ground truth. While F-mining and wavelets slightly improve the reconstruction, the spectral discriminator decreases image fidelity.
  }
  \label{fig:downsampling_artifacts}
  \vspace{-0.3cm}
\end{figure}

\boldparagraphwocolon{How does downsampling affect the training signal?}
\begin{figure}[t!]
  \centering
  \begin{minipage}{0.47\linewidth}
    \centering
    \begin{subfigure}[b]{0.31\linewidth}
   	 {\raisebox{0.5cm}{\rotatebox[origin=l]{90}{\small{Toyset}}}}
   	  \includegraphics[width=0.811\linewidth]{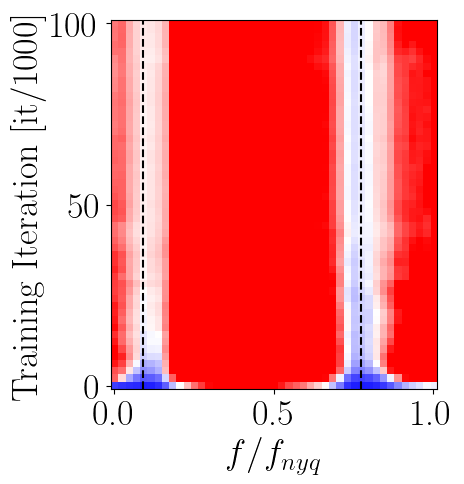}\\
   	  {\raisebox{0.4cm}{\rotatebox[origin=l]{90}{\small{CelebA}}}}
   	   \includegraphics[width=0.811\linewidth]{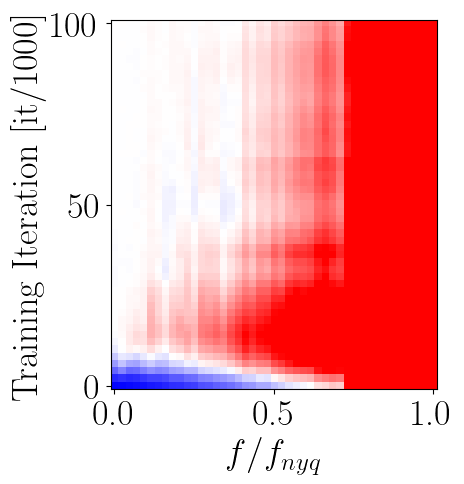}
   	   \caption{AvgPool}
   	   \label{fig:discriminator_avg}
   	\end{subfigure}
   	\begin{subfigure}[b]{0.208\linewidth}
      \includegraphics[width=\linewidth]{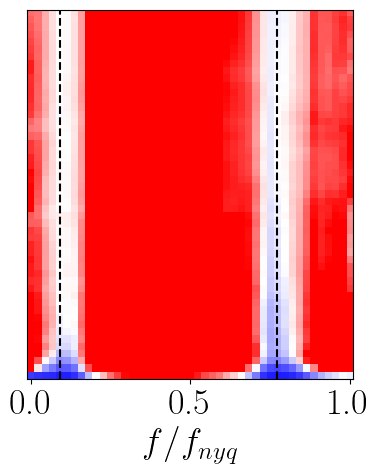}\\
      \includegraphics[width=\linewidth]{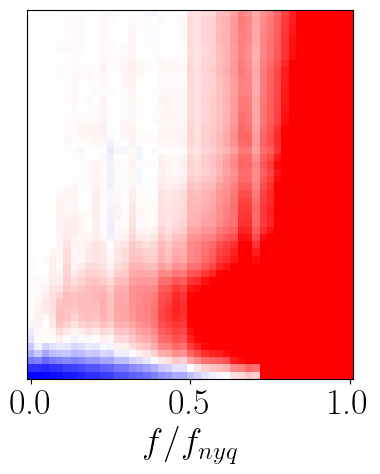}
         	   \caption{\scalebox{0.82}[1]{BlurPool}}
   	   \label{fig:discriminator_blurpool}
   	\end{subfigure}
   	\begin{subfigure}[b]{0.208\linewidth}      
      \includegraphics[width=\linewidth]{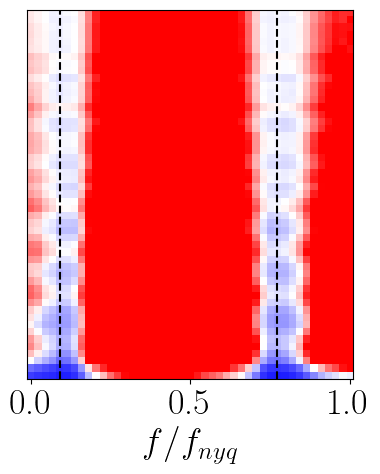}
      \includegraphics[width=\linewidth]{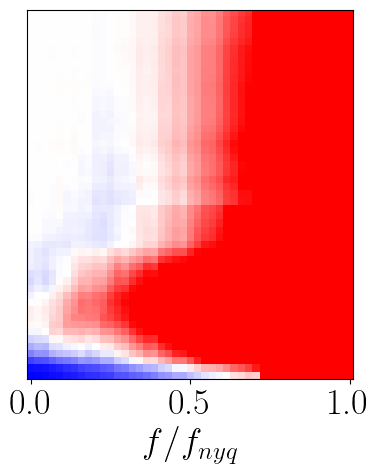}     
         	   \caption{Stride}
   	   \label{fig:discriminator_stride} 
   	\end{subfigure}
   	\begin{subfigure}[b]{0.237\linewidth}
      \includegraphics[width=\linewidth]{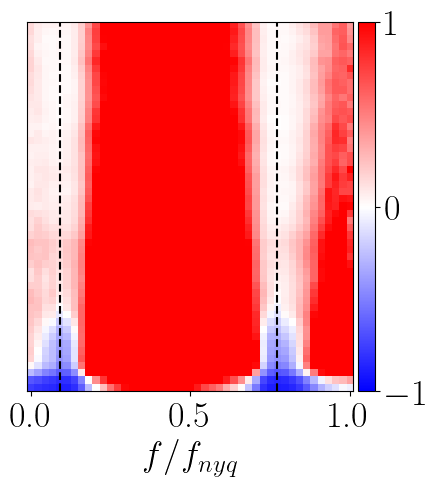}
      \includegraphics[width=\linewidth]{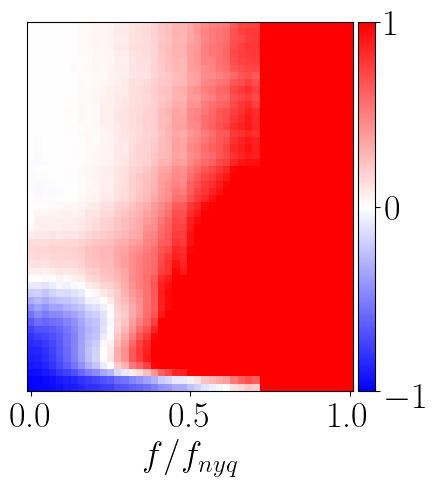}      
         	   \caption{MLP}
   	   \label{fig:discriminator_mlp}
	\end{subfigure}
  \caption{\textbf{Spectrum Error Evolution for Discriminators with Different Downsampling Operations. }
  The downsampling operations do not significantly bias the discriminator towards any frequency range. 
  Instead, it generally struggles with frequencies of low magnitude. 
  The color corresponds to the relative error of the average predicted reduced spectrum \wrt the ground truth and is clipped at $1$, i.e. when the relative error exceeds $100\%$.
  }
  \label{fig:discriminator}
  \end{minipage}\hfill%
  \begin{minipage}{0.47\linewidth}
    \centering
    \begin{subfigure}[b]{0.31\linewidth}
   	 {\raisebox{0.5cm}{\rotatebox[origin=l]{90}{\small{Toyset}}}}
   	  \includegraphics[width=0.811\linewidth]{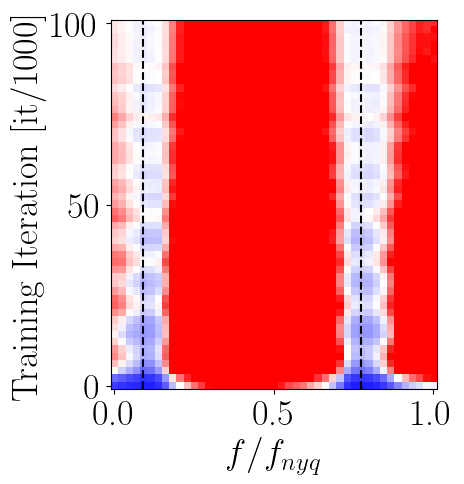}\\
   	  {\raisebox{0.4cm}{\rotatebox[origin=l]{90}{\small{CelebA}}}}
   	   \includegraphics[width=0.811\linewidth]{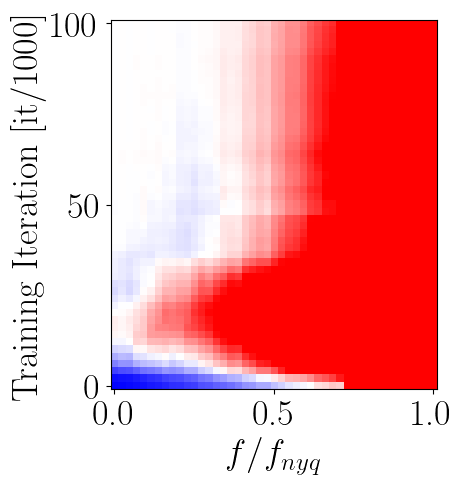}
   	   \caption{Stride}
   	   \label{fig:discriminator_domain_stride}
   	\end{subfigure}
   	\begin{subfigure}[b]{0.208\linewidth}
      \includegraphics[width=\linewidth]{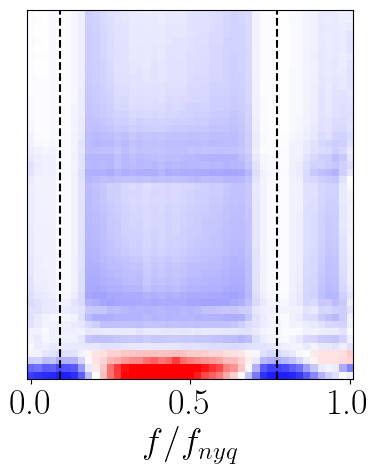}\\
      \includegraphics[width=\linewidth]{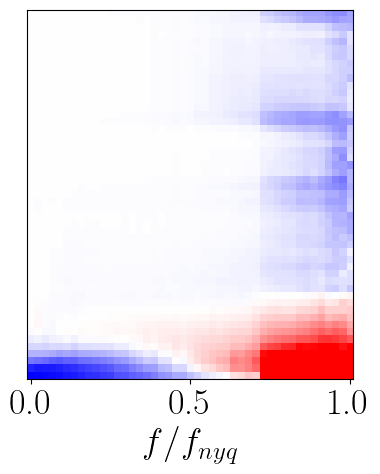}
       \caption{+SD}
   	   \label{fig:discriminator_domain_sd}
   	\end{subfigure}
   	\begin{subfigure}[b]{0.208\linewidth}      
      \includegraphics[width=\linewidth]{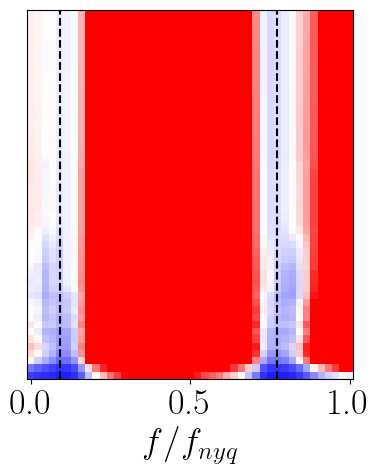}
      \includegraphics[width=\linewidth]{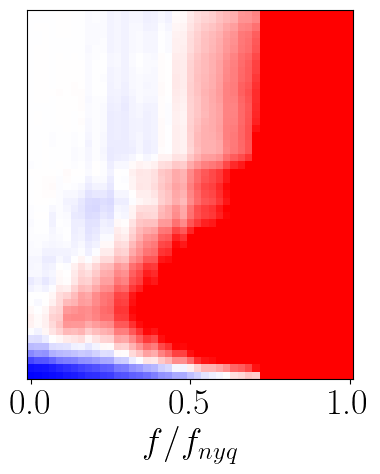}     
         	   \caption{\scalebox{0.66}[1]{+F-Mining}}
   	   \label{fig:discriminator_domain_fmining} 
   	\end{subfigure}
   	\begin{subfigure}[b]{0.237\linewidth}
      \includegraphics[width=\linewidth]{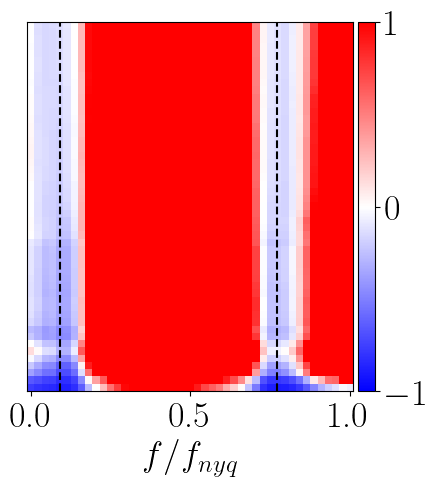}
      \includegraphics[width=\linewidth]{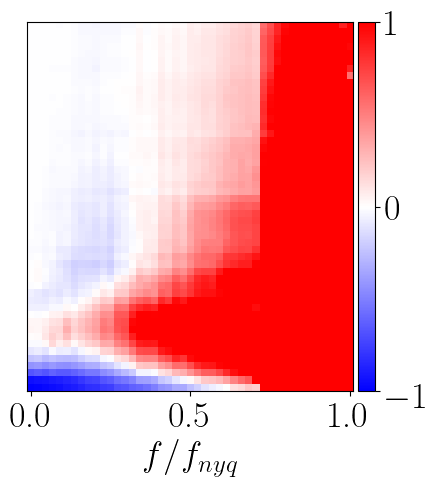}      
         	   \caption{\scalebox{0.9}[1]{+Wavelet}}
   	   \label{fig:discriminator_domain_wavelet}
	\end{subfigure}
  \caption{\textbf{Spectrum Error Evolution for Discriminators on Different Input Domains.}
  The spectral discriminator greatly improves the spectral statistics on both datasets while hard example mining in the frequency domain (F-mining) and wavelets alter results only slightly. 
  The color corresponds to the relative error of the average predicted reduced spectrum \wrt the ground truth and is clipped at $1$, i.e. when the relative error exceeds $100\%$.}
  \label{fig:discriminator_domain}
  \end{minipage}\hfill%
   \vspace{-0.3cm}
\end{figure}
\figref{fig:discriminator} shows the evolution of the average spectrum of the reconstructed images. 
On the Toyset, both the low- and high-frequency peaks are learned approximately at an equal pace. 
This suggests that the discriminator can indeed detect and correct the spectral statistics at high frequencies, regardless of the choice of downsampling operation. 
However, all approaches struggle to correct frequencies of low magnitude. 
For natural images, the power spectrum decays exponentially which creates the impression that the discriminator struggles with high frequencies, while it actually struggles with the low magnitude of the high-frequency content.\\
In image space, amongst the downsampling operations strided convolution achieves the best PSNR.
However, it is much lower than the values for the generator in \tabref{tab:generator} which reflects the harder task due to instance- instead of pixel-level supervision.
\figref{fig:downsampling_artifacts} shows the learned tensors at the last training iteration. While the reconstruction from the MLP is reasonably good, the downsampling operations introduce artifacts in the training signal provided by the discriminator.

\boldparagraphwocolon{Is it enough to train in the spatial domain?}
In \figref{fig:discriminator_domain} we compare the spectrum evolution for an additional spectral discriminator (SD)~\cite{Jung2021AAAI}, hard example mining in the frequency domain (F-Mining)~\cite{Chen2021AAAI} and training in wavelet space  (Wavelet)~\cite{Gal2021ARXIV}.
Since the spectral discriminator is directly applied to the reduced spectrum, it can guide the generator on the spectral statistics of the dataset and greatly reduces the error in~\figref{fig:discriminator_domain_sd}.
Instead, F-mining and wavelets only slightly reduce the spectral discrepancies, e.g., on CelebA for frequencies between $0.5-0.75f_{nyq}$.
While F-mining increases the weight of samples with poor spectral realness, it does not make the discriminator more sensitive to slight changes in image space. This is in agreement with the findings in~\cite{Chen2021AAAI}.
Wavelets separate the input \wrt the frequency but not \wrt the magnitude which could explain why low magnitude artifacts remain difficult to detect. \\
In the spatial domain, however, the reconstructions with the spectral discriminator  in \figref{fig:downsampling_artifacts_stride} and~\ref{fig:downsampling_artifacts_sd} reveal a caveat: While the spectrum is better aligned, the reconstruction with the additional spectral discriminator qualitatively becomes worse.
As the spectrum computation and the azimuthal integration discard information, images with the same reduced spectra can look very different. Consequently, penalizing the reduced spectrum might not be sufficient for improving the training signal alone.
This also reflects in the PSNR in \tabref{tab:discriminator} which, except for the MLP, remains largely unaffected by the spectral discriminator.
In the supplementary, we verify that replacing the reduced spectrum with the full Fourier transform indeed improves the image fidelity. However, as the spectral discriminator is an MLP, this does not trivially scale to real-world settings.

\boldparagraph{Implications}
In contrast to existing hypotheses, our findings evidence that the discriminator generally struggles with frequencies that have a low magnitude but that high frequencies are not per se more difficult to detect.
An additional discriminator on the reduced spectrum can greatly facilitate learning the spectral statistics of the data but might not improve the image fidelity. 
Even in our simple testbed, none of the convolutional architectures with downsampling is able to provide artifact-free supervision.
This indicates that the discriminator might play a more important role in reducing high-frequency artifacts than currently anticipated in the field. While a simple spectral discriminator provides a good first step, we conclude that more work in this area is required to solve the problem. 
We believe that one key is to reduce the downsampling artifacts, \eg, by exploring alternative downsampling operations or by considering approaches that allow for pixel-level supervision of the discriminator as in \cite{Shocher2020ARXIV} or \cite{Schonfeld2020CVPR}.

%% file: sec_gans.tex
\section{Improving GANs}\label{sec:gans}
We now analyze the full GAN training, \ie, when generator and discriminator are two competing neural networks. 
First, we extend the discriminator analysis from~\secref{sec:discriminator} and verify if these results are also valid for full GAN training.
Next, we investigate if the most effective discriminator is able to resolve high-frequency artifacts of the different upsampling strategies discussed in~\secref{sec:generator}.
Lastly, we analyze its effect on StyleGAN2~\cite{Karras2020CVPR}, a state-of-the-art GAN model, which shows the characteristic peak at the highest frequencies in the reduced spectrum.

\boldparagraph{Experimental Setting}
We start by combining the architectures from~\secref{sec:generator} and \secref{sec:discriminator} to obtain PGAN~\cite{Karras2018ICLR} with a reduced number of channels, which we train from scratch with R1-regularization and without progressive growing~\cite{Karras2020CVPR,Mescheder2018ICML} until the discriminator has seen $15$M images.
Next, we extend our analysis to StyleGAN2 on resolutions up to $1024^2$ pixels.
The StyleGAN2 generator uses bilinear upsampling which, according to our previous findings, should not cause the elevated amount of high frequencies. Therefore, we finetune pre-trained models with the most effective discriminator
following the training protocol from~\cite{Karras2020NIPS} until the discriminator has seen $2.5$M images. 

\boldparagraph{Datasets}
We consider large-scale real-world datasets in this section. We train our version of PGAN on a downsampled version of FFHQ~\cite{Karras2019CVPR} at resolution $64^2$ pixels and a downsampled version of 200k images from LSUN Cats~\cite{Yu2015ARXIV} at resolution $128^2$ pixels.
We finetune StyleGAN2 on LSUN Cats ($256^2$ pixels), AFHQ Dog~\cite{Choi2020CVPR} ($512^2$ pixels) and FFHQ ($1024^2$ pixels). For AFHQ Dog, we use adaptive discriminator augmentation due to the small size of the dataset~\cite{Karras2020NIPS}.

\boldparagraph{Evaluation Metrics}
In the spatial domain, we report FID~\cite{Heusel2017NIPS} on the full dataset and 50k generated images. To measure spectral discrepancies we deploy the spectral classifier described in \secref{sec:preliminaries}.
We follow the proposal in~\cite{Chandrasegaran2021CVPR} and replace the KNN classifier with an SVM to obtain a stronger classifier. We use $1000$ real and fake images each and use $90\%$ and $10\%$ for training and evaluation, respectively.
Here, a classification accuracy of around $50\%$ is ideal as this indicates a perfect generator because the classifier is forced to make a random guess.
\begin{figure}[t!]
    \centering
    \begin{subfigure}{0.222\linewidth}
    	\includegraphics[width=\linewidth]{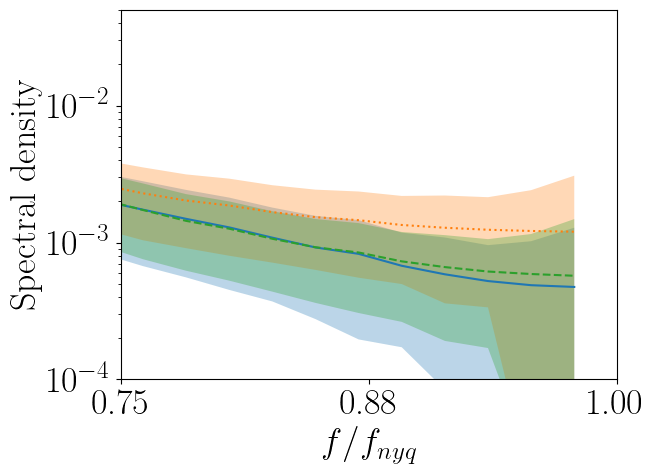}
    	\caption{Bilinear}
    \end{subfigure}
    \begin{subfigure}{0.2\linewidth}
    	\includegraphics[width=\linewidth]{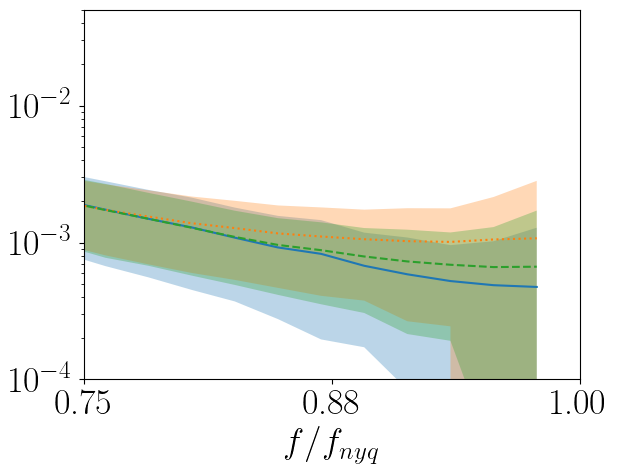}
    	\caption{NN}
    \end{subfigure}
    \begin{subfigure}{0.2\linewidth}
    	\includegraphics[width=\linewidth]{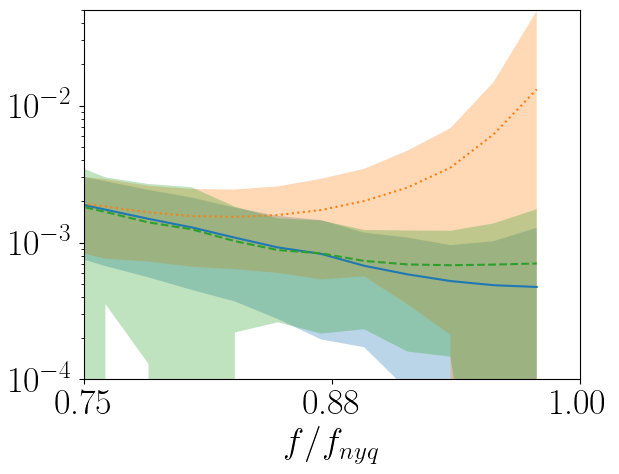}
    	\caption{Zeros}
    \end{subfigure}
    \begin{subfigure}{0.2\linewidth}
    	\includegraphics[width=\linewidth]{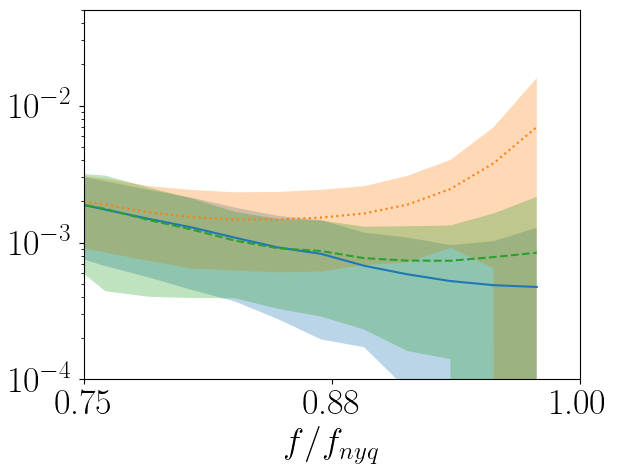}
    	\caption{Reshape}
    \end{subfigure}
    \begin{subfigure}[b]{0.14\linewidth}
    	\includegraphics[width=\linewidth]{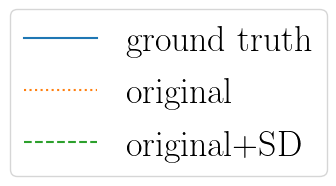}
    \end{subfigure}
    \caption{\textbf{Reduced Spectrum for PGAN} on FFHQ64. We plot the mean and standard
deviation of the reduced spectrum above $0.75f_{nyq}$.
     The spectral discriminator prevents the high peak for zero insertion and reshaping but cannot fully correct the spectral discrepancies. Similar to~\figref{fig:generator_specloss}, bilinear upsampling struggles to match the spectral statistics of the dataset.}
  \label{fig:pggan_upsampling_spectra}
    \vspace{-0.3cm}
\end{figure}
\begin{table}
\centering
		\begin{minipage}{0.48\textwidth}
	\centering
  		\setlength\tabcolsep{0.2em}
		\renewcommand{\arraystretch}{1.0}
  		\begin{small}
      \scalebox{0.9}{
  		\input{tab/gans/baselines_clean.tex}
      }
  		\end{small}
  		\vspace{0.1cm}
   		\caption{\textbf{Spectral Classification Accuracy and FID} for PGAN with discriminators on different input domains.
      }
    	\label{tab:baselines}
    \end{minipage}\hfill
	\begin{minipage}{0.48\textwidth}
	\centering
  		\setlength\tabcolsep{0.2em}
		\renewcommand{\arraystretch}{1.0}
  		\begin{small}
      \scalebox{0.9}{
  		\input{tab/gans/pggan_clean.tex}
      }
  		\end{small}
  		\vspace{0.1cm}
   		\caption{\textbf{Spectral Classification Accuracy and FID} for PGAN using different upsampling strategies with spectral discriminator.}
    	\label{tab:pggan_upsampling}
    \end{minipage}\hfill
	  \vspace{-0.5cm}
\end{table}
\begin{wraptable}{r}{0.34\textwidth}
\centering
  \begin{minipage}{0.34\textwidth}
  \centering
      \setlength\tabcolsep{0.2em}
    \renewcommand{\arraystretch}{1.0}
      \begin{small}
      \scalebox{0.9}{
      \input{tab/gans/stylegan2_clean.tex}}
      \end{small}
      \vspace{0.1cm}
      \caption{\textbf{Finetuning StyleGAN2} with an additional disciminator on the reduced spectrum.}
      \label{tab:stylegan2}
    \end{minipage}\hfill
\vspace{-0.4cm}
\end{wraptable}

\boldparagraphwocolon{How do the isolated settings transfer to full GAN training?}
We ablate PGAN with discriminators on different input domains in~\tabref{tab:baselines}. 
In agreement with the findings from~\secref{sec:discriminator}, the wavelet discriminator (Wavelet) and hard frequency mining (F-Mining) improve the spectral statistics only slightly. Hence, generated images can still be classified with high accuracy. The most effective method to learn the spectral statistics remains the additional spectral discriminator (SD) as indicated by the lower accuracy of the spectral classifier on all datasets. Consistent with our observation on the testbed, the image quality in the spatial domain remains largely unaffected.\\
Recalling the implication of \secref{sec:generator}, that the generator can learn to compensate for high-frequencies artifacts given a suited training objective, we now investigate whether the spectral discriminator satisfies such a requirement for different upsampling operations in the generator.  
Consistent with our observation on the generator testbed, \figref{fig:pggan_upsampling_spectra} shows that the spectrum discriminator is also able to significantly reduce the peak at the highest frequency for both zero insertion and reshaping upsampling. However, the magnitude at the highest frequencies remains slightly elevated because the generator only receives supervision through the discriminator (real vs. fake) instead of full ground truth spectra considered in the testbed. 
On the other hand, the bias towards little high-frequency content for bilinear and nearest neighbor upsampling aligns well with the spectral statistics of the datasets.
This also reflects in~\tabref{tab:pggan_upsampling} where images generated with zero insertion and reshaping are still detected with higher accuracy than images generated with bilinear and nearest neighbor upsampling. 
Considering both the spatial statistics and image fidelity, we observe that upsampling with nearest neighbor yields the best performance.

Lastly, we evaluate the effect of the spectral discriminator when applied to StyleGAN2 on datasets with higher resolution, namely LSUN Cats ($256^2$), AFHQ Dog ($512^2$) and FFHQ ($1024^2$). The results in~\tabref{tab:stylegan2} show mixed results: On AFHQ Dog the spectral discriminator results in a strong improvement on the spectrum but also significantly increases the FID. In contrast, on FFHQ the FID stays similar but the spectral discriminator also only slightly improves the spectral statistics and is not able to correct the peak at the highest frequency\footnote{Jung \etal~\cite{Jung2021AAAI} conduct a similar experiment but evaluate the average over the logarithmic reduced spectra, \ie, $\frac{1}{N}\sum_{i=1}^N \log(\tilde{S}_i)$ while we consider $\frac{1}{N}\log\left(\sum_{i=1}^N \tilde{S}_i\right)$.  We find that the peak at the highest frequencies is only significant in the latter computation and confirmed this difference with the authors.}.
This suggests that in existing architectures there is a tradeoff between perceptual image quality and matching the spectral statistics of the data. 

\boldparagraph{Implications}
Our experiments show that both of our testbeds on the generator and the discriminator provide consistent observations which are aligned with those for full GAN training.
Our findings suggest that the spectral discriminator can reduce the spectral discrepancies but the misalignment in the high frequencies is not fully addressed.
We further observe that nearest neighbor upsampling together with an additional spectral discriminator is a good combination for natural images where high-frequency content is low.
However, when applying existing measures to StyleGAN2, we observe that none of them can solve the spectral discrepancies completely.
This suggests that the high-frequency artifacts in GANs are still an open problem that require further research. Our work takes an important step towards understanding the underlying mechanisms and sheds light on the effectiveness and limitations of existing approaches.

%% file: tab/gans/baselines_clean.tex
\begin{tabular}{lcccccccc}
\toprule
 & \multicolumn{2}{c}{Original} & \multicolumn{2}{c}{Wavelet} & \multicolumn{2}{c}{F-Mining} & \multicolumn{2}{c}{SD} \\
 			& Acc 		& FID 			& Acc 		& FID 		& Acc 		& FID 		& Acc 		& FID 		\\
\midrule
FFHQ64 		&   $72$ 	& 31.0 		& $ 73$ 	& 	39.8 	& $73$ 	& 	33.7		& $ 62$  	& 	31.0 	\\
Cats128 	& 	$88$ 	& 106.1 	& $ 74$ 	& 	122.5 	& $77$  & 	119.3	& $ 62$		& 	114.6 	\\

\bottomrule     
\end{tabular}

%% file: tab/gans/pggan_clean.tex
\begin{tabular}{lcccccccc}
\toprule
 & \multicolumn{2}{c}{Bilinear} & \multicolumn{2}{c}{NN} & \multicolumn{2}{c}{Zeros}  & \multicolumn{2}{c}{Reshape}\\
 & Acc & FID & Acc & FID & Acc & FID & Acc & FID \\
\midrule
FFHQ64 & 	$57$ 	&	36.3 	& 	$62$ &  31.0 	&	$ 66$	&	34.9	& 	$67$	& 	35.2  \\
Cats128 & 	$62$ 	&	138.2 	& 	$62$ &  114.6 	&	$ 77$	&	128.8	& 	$70$	& 	129.7  \\
\bottomrule     
\end{tabular}

%% file: tab/gans/stylegan2_clean.tex
\begin{tabular}{lcccc}
\toprule
 & \multicolumn{2}{c}{Original} & \multicolumn{2}{c}{+SD} \\
 & Acc & FID & Acc & FID \\
\midrule
Cats256  &	$92$		& 	6.5	&	$80$	&	8.4	\\
AFHQ Dog & 	$92$		& 	7.4 	& 	$66$	& 	12.1 \\
FFHQ 	 & 	$99$		&  	2.8  	& 	$94$	&	3.1  \\
\bottomrule     
\end{tabular}

%% file: sec_related.tex
\section{Other Related Work}

\boldparagraph{Deepfake Detection}
In the last years, great progress on unconditional Generative Adversarial Networks enabled photo-realistic image synthesis with deep neural networks \cite{Goodfellow2014NIPS,Karras2020CVPR,Karras2019CVPR,Karras2018ICLR,Mescheder2018ICML,Anokhin2021CVPR,Skorokhodov2021CVPR}. 
With the rapid development in image synthesis, fake image detection becomes equally important~\cite{MarraGCV2018,Roessler2019ICCV,Wang2020CVPR}. 
Albeit very high photorealism, multiple works show that generated images can still be easily distinguished from real images based on their high-frequency content \cite{Jiang2020ARXIV,Wang2020CVPR,Dzanic2020NIPS,Durall2020CVPR,LiuZ2020CVPR,Zhang2019IEEE,Frank2020ICML}. 
Consistent behavior across various architectures suggests a systematic problem in state-of-the-art GANs. Thus, it is important to understand if there is a frequency bias in convolutional generators or discriminators.

\boldparagraph{Spectral Bias}
Fully connected ReLU networks are known to fit low-frequency modes of a target function faster than its high-frequency modes~\cite{Jacot2018NIPS,Rahaman2019ICML,Xu2019ARXIV,Arora2019ICML,Basri2019NIPS}.
This behavior, referred to as \textit{spectral bias}, has been shown to greatly impact the practical expressiveness of such models due to frequency dependent learning rates~\cite{Jacot2018NIPS,Rahaman2019ICML,Xu2019ARXIV,Arora2019ICML,Basri2019NIPS}. 
For CNNs, a spectral bias is often mentioned to explain the good generalization of largely overparameterized networks~\cite{WangH2020CVPR,Ulyanov2018CVPR,Xu2019ARXIV} but is theoretically less well understood. 
Inspired by Rahaman \etal~\cite{Rahaman2019ICML} who study the spectral bias of fully connected ReLU networks, we analyze if a similar bias exists in convolutional generators and discriminators. Note that Rahaman \etal~\cite{Rahaman2019ICML} consider one-dimensional regression tasks in which the spectrum is evaluated over the \textit{output samples}. Instead, we calculate the spectrum over \textit{all pixels} of one image. To highlight this difference, we use the term \textit{frequency bias} in our work instead.

%% file: sec_conclusion.tex
\section{Conclusion}\label{sec:conclusion}
In this work, we take a thorough look at existing explanations for systematic artifacts in the spectral statistics of generated images and unify the efforts that have been done so far.
While our experiments suggest that upsampling introduces a frequency bias in the generator
this alone does not explain the spectral discrepancies. 
In agreement with~\cite{Jung2021AAAI,Chen2021AAAI}, we find that advancing the discriminator brings us one step closer to learning the correct data distribution. 
However, none of the existing measures can faithfully recover the spectral statistics yet, such that generated images are still straightforward to classify solely based on their spectra. 
While state-of-the-art GANs benefit from highly engineered generator architectures, \eg,~\cite{Karras2018ICLR,Karras2019CVPR,Karras2020CVPR}, our findings suggest that the design of the discriminator plays an equally important role and deserves more attention in future work.
We encourage future work to explore alternative downsampling operations and to consider approaches that enable pixel-level supervision for the discriminator. Interesting starting points could for example be \cite{Shocher2020ARXIV} or \cite{Schonfeld2020CVPR}. Another interesting direction might be to explore discriminator augmentation~\cite{Karras2020NIPS,Zhao2020NIPS} \wrt high-frequency artifacts.
We believe that our testbeds for both the generator and discriminator may serve as useful tools for future investigations and will make our code publicly available upon acceptance.
Lastly, we remark that working on generative models always bears the risk of manipulation and the creation of misleading content. While the insights from our work could be used to make Deepfakes harder to detect, we believe that a better understanding of existing architectures and their limitations is key to advancing the field, such that the benefits of our work outweigh the societal risks.

%% file: sec_acknowledgments.tex
\section*{Acknowledgments}
We acknowledge the financial support by the BMWi in the project KI Delta Learning (project number 19A19013O) and the support from the BMBF through the Tuebingen AI Center (FKZ: 01IS18039A). We thank the International Max Planck Research School for Intelligent Systems (IMPRS-IS) for supporting Katja Schwarz. This work was supported by an NVIDIA research gift.

%% file: supplementary/sec_testbeds.tex
\section{Implementation}\label{sec:testbeds}
\subsection{Generator}
The generator architecture is specified in~\tabref{sup:tab:generator} and is the same as the PGAN~\cite{Karras2018ICLR} generator except for a reduced number of channels. For the model, we base our implementation on~\url{https://github.com/rosinality/progressive-gan-pytorch}.
Following ~\cite{Karras2018ICLR}, we use a slope of $0.2$ for the LeakyRelu and apply pixel normalization after every convolution. The first convolution uses a padding of $3$, while the remaining convolutions retain the input resolution.\\
Note that upsampling by reshaping reduces the channel dimension by a factor of $4$. Therefore, in this case, we multiply the output channels $c_{out}$, \cf~\tabref{sup:tab:generator}, of each convolutional layer before the upsampling by $4$. To keep a similar amount of total parameters, we further divide all channel dimensions by $1.5$. \\
We train the model end-to-end using Adam optimizer~\cite{Kingma2015ICLR} with a learning rate of $0.0001$ and a batch size of $10$. 
When training with both an L2-loss in image space and an L2-loss on the reduced spectrum, we weigh both losses equally.
We train on a single NVIDIA Tesla V100-SXM2-32GB.\\
\boldparagraphnospace{Teaser}
For the experiment in Fig. 1b of the main paper, we train on a single image of resolution $128$. We use the settings mentioned before but use a batch size of $1$.
\begin{table}
    \centering
    \begin{small}
    \begin{tabular}{lllllll}
        \toprule
        Layer Type & Kernel Size & $c_{in}$ & $c_{out}$  & Activation & Normalization & Repetitions\\
        \midrule
        Conv & 4 & $64$ & $64$ & LeakyRelu  & PixelNorm & 1\\
        Conv & 3 & $64$ & $64$ & LeakyRelu  & PixelNorm & 1 \\
        Upsample & -- & $64$ & $64$ & --  & PixelNorm & \multirow{3}{*}{\hspace{-1em}$\left.\begin{array}{l}
                \\
                \\
                \\
                \end{array}\right\rbrace\times{n}$}\\
        Conv & 3 & $64$ & $64$ & LeakyRelu  & PixelNorm &  \\
        Conv & 3 & $64$ & $64$ & LeakyRelu  & PixelNorm &  \\
        Upsample & -- & $64$ & $64$ & --  & PixelNorm &  \multirow{3}{*}{\hspace{-1em}$\left.\begin{array}{l}
                \\
                \\
                \\
                \end{array}\right\rbrace\times{1}$}\\
        Conv & 3 & $64$ & $32$ & LeakyRelu  & PixelNorm &  \\
        Conv & 3 & $32$ & $32$ & LeakyRelu  & PixelNorm &  \\
        Upsample & -- & $32$ & $32$ & --   & PixelNorm & \multirow{3}{*}{\hspace{-1em}$\left.\begin{array}{l}
                \\
                \\
                \\
                \end{array}\right\rbrace\times{1}$}\\
        Conv & 3 & $32$ & $16$ & LeakyRelu  & PixelNorm &  \\
        Conv & 3 & $16$ & $16$ & LeakyRelu  & PixelNorm &  \\
	    Conv & 1 & $16$ & $3$ & -- & PixelNorm & 1 \\
	\bottomrule
    \end{tabular}
    \end{small}
    \vspace{0.2cm}
    \caption{\textbf{Architecture of the PGAN~\cite{Karras2018ICLR} Generator with Reduced Channels.} The value of $n$ depends on the resolution of the data, \eg, $n=2$ and $n=3$ for resolution $64^2$ and $128^2$ pixels, respectively.}
    \label{sup:tab:generator}
            \vspace{-0.3cm}
\end{table}

\subsection{Discriminator}
\begin{table}
    \centering
    \begin{small}
	\begin{tabular}{llllll}
        \toprule
        Layer Type & Kernel Size & $c_{in}$ & $c_{out}$  & Activation & Repetitions\\
        \midrule
        Conv & 1 & $3$ & $16$ & LeakyRelu  & 1\\
		Conv & 3 & $16$ & $32$ & LeakyRelu  & 1\\
        Conv & 3 & $32$ & $32$ & LeakyRelu & 1 \\
        Downsample & -- & $32$ & $32$ & -- & \multirow{3}{*}{\hspace{-1em}$\left.\begin{array}{l}
                \\
                \\
                \\
                \end{array}\right\rbrace\times{1}$}\\
        Conv & 3 & $32$ & $64$ & LeakyRelu &  \\
        Conv & 3 & $64$ & $64$ & LeakyRelu &  \\
        Downsample & -- & $64$ & $64$ & -- &  \multirow{3}{*}{\hspace{-1em}$\left.\begin{array}{l}
                \\
                \\
                \\
                \end{array}\right\rbrace\times{n}$}\\
        Conv & 3 & $64$ & $64$ & LeakyRelu &  \\
        Conv & 3 & $64$ & $64$ & LeakyRelu &  \\
        Downsample & -- & $64$ & $64$ & --  & \multirow{3}{*}{\hspace{-1em}$\left.\begin{array}{l}
                \\
                \\
                \\
                \end{array}\right\rbrace\times{1}$}\\
        Conv & 3 & $65$ & $64$ & LeakyRelu &  \\
        Conv & 4 & $64$ & $64$ & LeakyRelu &  \\
	    Linear & -- & $64$ & $n_c$ & --  & 1 \\
	\bottomrule
    \end{tabular}
    \end{small}
    \vspace{0.2cm}
    \caption{\textbf{Architecture of the PGAN~\cite{Karras2018ICLR} Discriminator with Reduced Channels.} $n_c$ denotes the number of classes. The value of $n$ depends on the resolution of the data, \eg, $n=2$ and $n=3$ for resolution $64^2$ and $128^2$ pixels, respectively.}
    \label{sup:tab:discriminator}
        \vspace{-0.3cm}
\end{table}
The discriminator architecture is specified in~\tabref{sup:tab:discriminator} and is the same as the PGAN~\cite{Karras2018ICLR} discriminator except for a reduced number of channels. For the model, we base our implementation on~\url{https://github.com/rosinality/progressive-gan-pytorch}. Following~\cite{Karras2018ICLR}, we use a slope of 0.2 for the LeakyRelu, and append the minibatch standard deviation after the last downsampling operation which increases $c_{in}$ in~\tabref{sup:tab:discriminator} from $64$ to $65$. The last convolution uses no padding to reduce the spatial size from $4\times4$ to $1\times1$ before the linear layer, while the remaining convolutions retain the spatial size. \\
For the reconstruction task, we train a class-conditional discriminator with a single sample per class. Hence, the linear layer outputs $n_c$ logits, where $n_c$ is the number of classes, \ie, samples. In our experiments, we use $n_c=10$.
The generator is replaced by $n_c$ learnable tensors which we optimize jointly with the discriminator. We train the model end-to-end using Adam optimizer~\cite{Kingma2015ICLR} with a learning rate of $0.0001$ for both the tensors and the discriminator, and a batch size of $10$. 
To stabilize training we train with R1-regularization using a regularization strength of $10$ and use an exponential moving average with decay $0.999$ for the learnable tensors to produce the images. For all of our experiments, we ensure that training is stable by verifying that the discriminator converges to equilibrium, \ie, that the logits of the discriminator approach zero during training. 
For our testbed, we build on the framework from \url{https://github.com/LMescheder/GAN_stability.git} because it is intuitive and straightforward to work with.
We train on a single NVIDIA Tesla V100-SXM2-32GB.\\
\boldparagraphnospace{Teaser}
For the experiment in Fig. 1c of the main paper, we train on a single image of resolution $128$. We use the settings mentioned before but use a batch size of $1$.

\subsection{GAN}
Our code framework for training the full GAN setting is based on the publicly available code for StyleGAN2-ada~\cite{Karras2020NIPS}:~\url{https://github.com/NVlabs/stylegan2-ada-pytorch.git} because it is optimized for large-scale multi-GPU GAN training. 
We train PGAN on two NVIDIA Tesla V100-SXM2-32GB GPUs with a batch size of $256$ and a learning rate of $0.002$ for both the generator and the discriminator. The strength of the $R1$-regularizer is $10$ and we use a minibatch size of $4$ to compute the standard deviation within the batches.
We do not train PGAN with adaptive discriminator augmentation and adhere to the training protocol from~\cite{Karras2020CVPR}.\\
For finetuning StyleGAN2 we follow the training process of~\cite{Karras2020CVPR} using four V100-SXM2-32GB or four GEFORCE GTX 1080 Ti.

%% file: supplementary/sec_generator.tex
\section{Additional Analysis of the Generator Testbed}\label{sup:sec:generator}
\subsection{Number of Images}
\begin{figure}[t!]
  \centering
  \begin{minipage}{0.47\linewidth}
    \centering
   	\includegraphics[width=\linewidth]{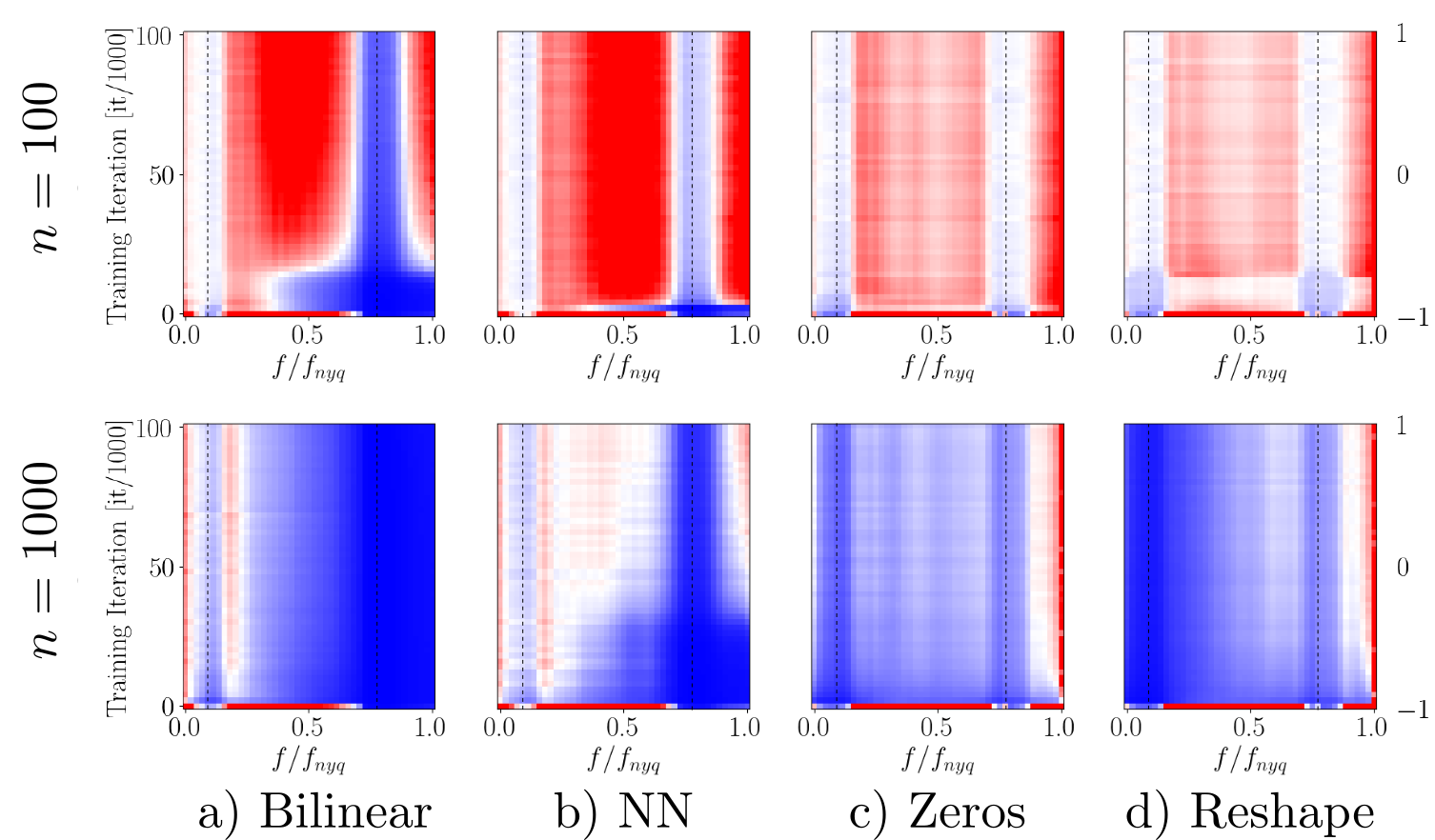}
  \caption{\textbf{Spectrum Error Evolution for Generators with Different Number of Training Samples on the Toyset. }
  Regardless of the number of training images, interpolation-based upsampling results in too little high-frequency content while zero insertion and reshaping struggle with checkerboard artifacts.
  The color corresponds to the relative error of the average predicted reduced spectrum \wrt the ground truth and is clipped at $1$, i.e. when the relative error exceeds $100\%$.
  }
  \label{fig:generator_ablate_n}
  \end{minipage}\hfill%
  \begin{minipage}{0.47\linewidth}
    \centering
   	  \includegraphics[width=\linewidth]{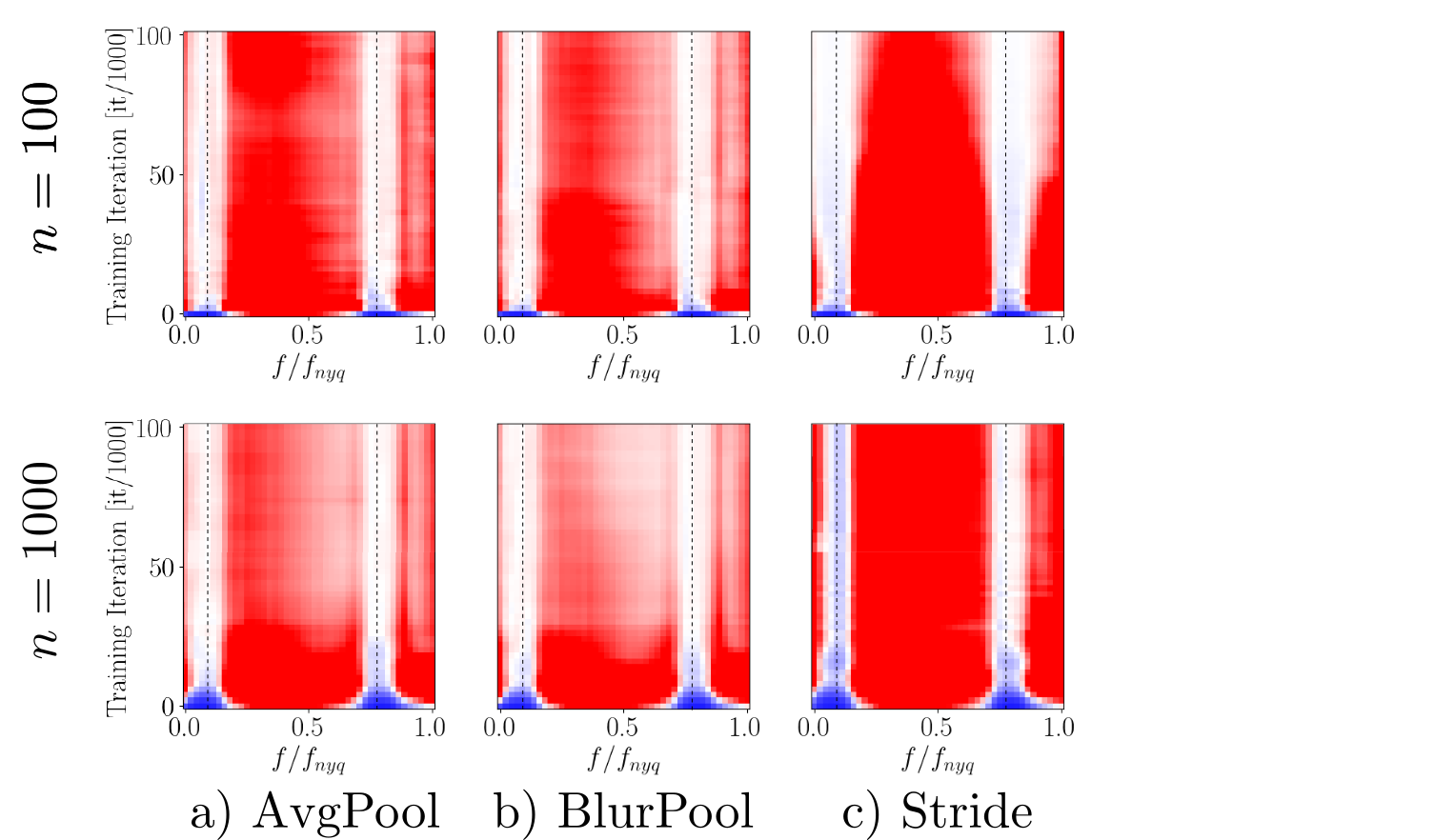}\\
  \caption{\textbf{Spectrum Error Evolution for Discriminators with Different Number of Training Samples on the Toyset. }
 Regardless of the number of training images, the discriminator shows no significant bias towards any frequency range. Instead, it generally struggles with frequencies of low magnitude.
  The color corresponds to the relative error of the average predicted reduced spectrum \wrt the ground truth and is clipped at $1$, i.e. when the relative error exceeds $100\%$.}
  \label{fig:disc_ablate_n}
  \end{minipage}\hfill%
   \vspace{-0.3cm}
\end{figure}
We ablate how the number of training images impacts our generator testbed. \figref{fig:generator_ablate_n} shows that the findings from the main paper remain consistent with a varying number of training images. 

\subsection{Low Magnitude Errors}
In this section, we derive that an L2-loss penalizes errors with low magnitudes in the frequency domain less than errors with high magnitudes.
Let us first consider a 1D-signal with discrete values $x_n$, $n=0,\dots, N-1$. According to Parseval's theorem, for the discrete Fourier transform $\mathcal F$ it holds that
\begin{equation}
	\sum_{n=0}^{N-1} |x_n|^2 = \frac{1}{N}\sum_{k=0}^{N-1} |\mathcal F[\bx]_k|^2
\end{equation}
where $|\mathcal F[\bx]_k|^2$ is the magnitude at frequency $k$.
For an L2-loss on a predicted signal with values $x^{pred}_n$ and a target signal with values $x^{tgt}_n$ we obtain
\begin{eqnarray}
	\sum_{n=0}^{N-1} \left|x^{pred}_n - x^{tgt}_n\right|^2 &=& \sum_{n=0}^{N-1} \left|\left[\bx^{pred} - \bx^{tgt}\right]_n\right|^2 \\
	&=& \frac{1}{N}\sum_{k=0}^{N-1} \left|\mathcal F \left[\bx^{pred} - \bx^{tgt}\right]_k\right|^2 \label{eq:parseval}
\end{eqnarray}
Let us now consider some frequency $k_0$. When the error at $k_0$ has a low magnitude, then $|\mathcal F \left[\bx^{pred} - \bx^{tgt}\right]_{k_0}|^2$ is small and therefore contributes only slightly to the sum in~Eq.~\eqref{eq:parseval}. \\
For a gray scale image with pixel values $x_{ij}$, $i=0,\dots, H-1$, $j=0,\dots, W-1$, this follows similarly from Parseval's theorem in 2D
\begin{equation}
\sum_{i=0}^{H-1}\sum_{j=0}^{W-1} |x_{ij}|^2 = \frac{1}{HW}\sum_{k=0}^{H-1}\sum_{l=0}^{W-1} |\mathcal F[\bx]_{kl}|^2
\end{equation}

%% file: supplementary/sec_discriminator.tex
\section{Additional Analysis of the Discriminator Testbed}\label{sup:sec:discriminator}
\subsection{Number of Images}
We ablate how the number of training images impacts our discriminator testbed. \figref{fig:disc_ablate_n} shows that the findings from the main paper remain consistent with a varying number of training images. 

\subsection{Penalizing the Spectrum}
Jung \etal~\cite{Jung2021AAAI} propose an additional discriminator on the reduced spectrum (SD). However, penalizing the reduced spectrum might not be sufficient for improving the training signal alone because the spectrum computation and the azimuthal
integration discard information, see Section 4 of the main paper. Therefore, we investigate if using the full Fourier transform instead of the reduced spectrum can improve performance. 
We compare two different additional discriminators to the version from~\cite{Jung2021AAAI}: The first is an MLP that operates on the flattened 2D Fourier transform of the image and therefore uses neither convolutions nor downsampling (SD-FT). Since this is only feasible for images with a small resolution, we also investigate a convolutional spectral discriminator (SD-CNN-FT) with the same architecture as the spatial discriminator. 
We weigh the spatial and spectral discriminator equally as done in~\cite{Jung2021AAAI}. Further, we ensure that the models of all approaches have a similar number of parameters. Note that using an MLP on the flattened 2D Fourier transform increases the number of parameters from $\sim300$k to $\sim900$k. Hence, we increase the channel dimensions of the discriminator for both SD and SD-CNN-FT to obtain models of comparable size. \\
The PSNR values in~\tabref{tab:discriminator_psnr} show a significant improvement for all downsampling techniques for SD-FT. However, with the convolutional variant, SD-CNN-FT, the PSNR does not improve \wrt SD. 
\begin{table}
\centering
	\begin{minipage}{0.46\textwidth}
	\centering
  		\setlength\tabcolsep{0.2em}
		\renewcommand{\arraystretch}{1.0}
  		\begin{small}
  		\scalebox{0.9}{
  		\input{tab/discriminator/psnr_supp.tex}}
  		\end{small}
  		\vspace{0.1cm}
   		\caption{\textbf{PSNR} for different spectral discriminators on CelebA at resolution $64^2$ pixels.}
    	\label{tab:discriminator_psnr}
    \end{minipage}\hfill	%
	\begin{minipage}{0.46\textwidth}
	\centering
  		\setlength\tabcolsep{0.2em}
		\renewcommand{\arraystretch}{1.0}
  		\begin{small}
 		\scalebox{0.9}{
 		\input{tab/gans/sngan.tex}}
  		\end{small}
  		\vspace{0.1cm}
   		\caption{\textbf{Spectral Classification Accuracy and FID} for SNGAN on Cats128 with discriminators on different input domains.}
    	\label{tab:discriminator_sngan}
    \end{minipage}
\end{table}
\begin{figure}
	\centering
  \begin{subfigure}[t]{0.17\linewidth}
  \includegraphics[width=\linewidth]{gfx/generator/celeba_64/train_sample.png}
  \caption{GT}
  \end{subfigure}
  \begin{subfigure}[t]{0.17\linewidth}
  \includegraphics[width=\linewidth]{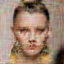}
  \caption{SD}  
  \end{subfigure}
  \begin{subfigure}[t]{0.17\linewidth}
  \includegraphics[width=\linewidth]{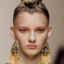}
  \caption{SD-FT}
  \end{subfigure}
  \begin{subfigure}[t]{0.17\linewidth}
  \includegraphics[width=\linewidth]{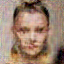}
   \caption{SD-CNN-FT}
   \end{subfigure}
   		\caption{\textbf{Reconstruction Guided by the Discriminator} with BlurPool downsampling for different spectral discriminators on CelebA at resolution $64^2$ pixels.}
    	\label{fig:discriminator_imgs}
  \vspace{-0.3cm}
\end{figure}
This is consistent with the qualitative results for BlurPool downsampling in~\figref{fig:discriminator_imgs}. 
Even with the increased amount of parameters, SD cannot remove the downsampling artifacts. While SD-FT closely reconstructs the ground truth, it can only be used with low-resolution images due to its fully connected architecture. Unfortunately, na\"ively applying a convolutional architecture on the full spectrum also suffers from artifacts in the reconstructions due to downsampling.
These observations suggest that integrating frequency domain supervision effectively and efficiently remains an open question in a GAN setting.

%% file: tab/discriminator/psnr_supp.tex
\begin{tabular}{lccc}
\toprule
 & AvgPool & BlurPool & Stride \\
\midrule
SD & 31.6 & 26.2 & 	27.7  \\
SD-FT & 53.9 & 53.0 & 	46.3 \\
SD-CNN-FT & 27.5 & 24.9 & 26.0 \\
\bottomrule     
\end{tabular}

%% file: tab/gans/sngan.tex
\begin{tabular}{lcccccccc}
\toprule
 & \multicolumn{2}{c}{Original} & \multicolumn{2}{c}{Wavelet} & \multicolumn{2}{c}{F-Mining} & \multicolumn{2}{c}{SD} \\
 			& Acc 		& FID 			& Acc 		& FID 		& Acc 		& FID 		& Acc 		& FID 		\\
\midrule
SNGAN   &   $75$    & 78.0      & $ 59$     &   59.2    & $65$  &   80.6    &  $58$    &   63.0       \\
\bottomrule     
\end{tabular}

%% file: supplementary/sec_gans.tex
\section{Additional Analysis of Full GAN Training}\label{sup:sec:gans}
\subsection{PGAN}
\begin{figure}[t!]
    \centering
    \begin{subfigure}{0.3\linewidth}
    	\includegraphics[width=\linewidth]{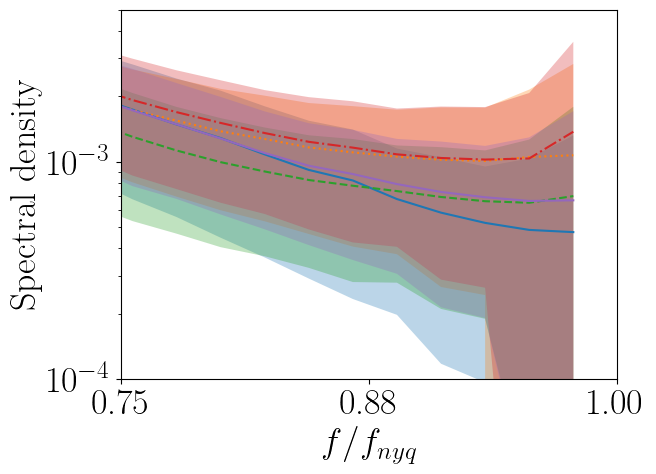}
    	\caption{FFHQ64}
    \end{subfigure}
    \hspace{1cm}
    \begin{subfigure}{0.3\linewidth}
    	\includegraphics[width=\linewidth]{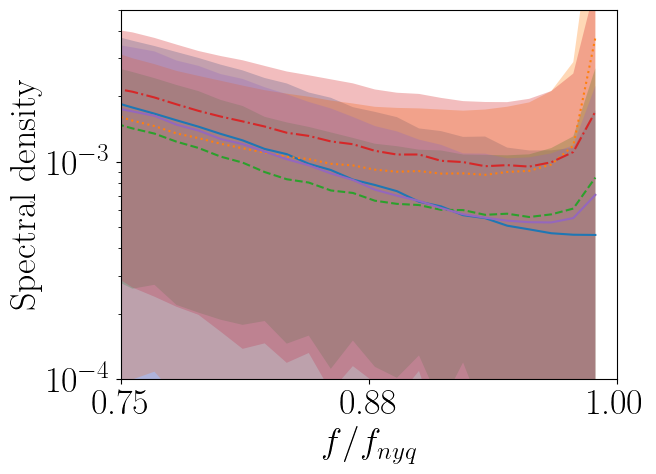}
    	\caption{Cats128}
    \end{subfigure}
    \raisebox{-0.15cm}{
    \includegraphics[width=0.17\linewidth]{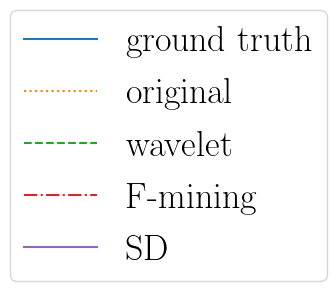}}
    \caption{\textbf{Reduced Spectrum for PGAN with Discriminators on Different Input Domains.} We plot the mean and standard
deviation of the reduced spectrum above $0.75f_{nyq}$.}
  \label{fig:pgan_discriminators}
    \vspace{-0.3cm}
\end{figure}
\boldparagraphnospace{Discriminators on Different Input Domains}
For our ablation on different input domains for the discriminator,~\figref{fig:pgan_discriminators} shows spectrum plots corresponding to Table 3 in the main paper. In agreement with the observations in~\cite{Chen2020ARXIV}, F-mining alters the spectral statistics at the highest frequencies only slightly. 
Wavelets reduce the peak at the highest frequencies but also predict too few frequencies below  $0.88f_{nyq}$. Similar to wavelets, the additional spectral discriminator (SD) reduces the peak at the highest frequencies but matches the spectral statistics below $0.88f_{nyq}$ more closely. These observations are consistent across both datasets and support the reported classification accuracy in Table 3 of the main paper.

\begin{figure}[t!]
    \centering
    \begin{subfigure}{0.4\linewidth}
    	\includegraphics[width=\linewidth]{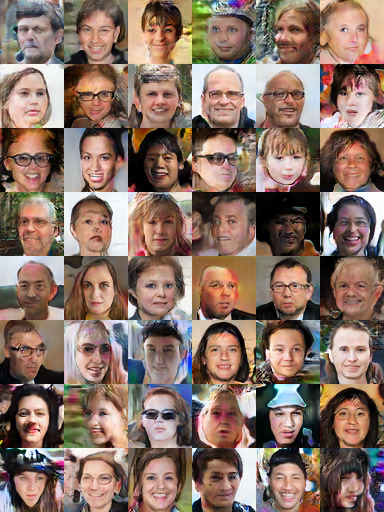}
    	\caption{FFHQ64}
    \end{subfigure}
    \hspace{1cm}
    \begin{subfigure}{0.4\linewidth}
    	\includegraphics[width=\linewidth]{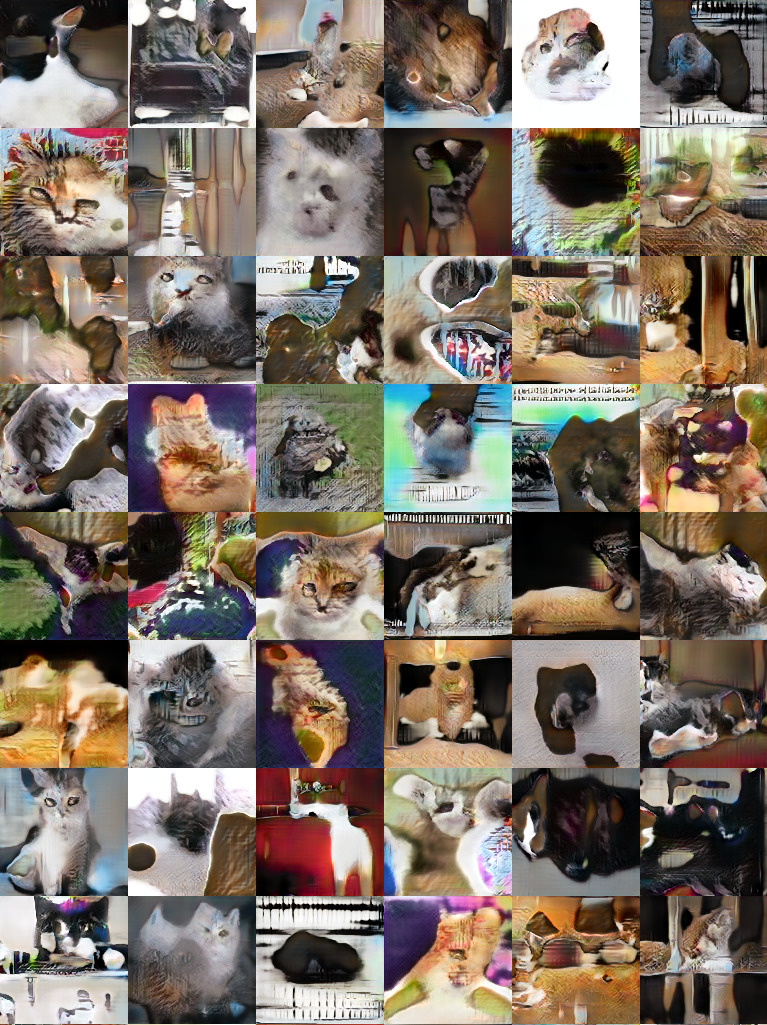}
    	\caption{Cats128}
    \end{subfigure}
    \caption{\textbf{Random Samples from PGAN with Reduced Channels.}}
  \label{fig:pgan_qualitative}
    \vspace{-0.3cm}
\end{figure}
\boldparagraphnospace{Qualitative Results for PGAN}
In~\figref{fig:pgan_qualitative}, we include some qualitative results for our smaller version of PGAN in the original setting. While not completely photo-realistic, the model can still reproduce characteristic features of the ground truth data. For our analysis, this is sufficient because the focus of these experiments is not on image fidelity but to study the spectral properties of both the generator and the discriminator in combination.
\subsection{SNGAN}
In this section, we consider different discriminators for SNGAN~\cite{Miyato2018ICLR} to investigate if our analysis from Section 4 in the main paper is consistent across architectures.
We base our framework on the official implementation of~\cite{Chen2021AAAI}, \url{https://github.com/cyq373/SSD-GAN.git}.
Similar to Section 4 in the main paper, we compare an additional spectral discriminator (SD)~\cite{Jung2021AAAI}, hard example mining in the frequency domain (F-Mining)~\cite{Chen2021AAAI}, and training in wavelet space (Wavelet)~\cite{Gal2021ARXIV}.
\begin{figure}[t!]
    \centering
    \includegraphics[width=0.3\linewidth]{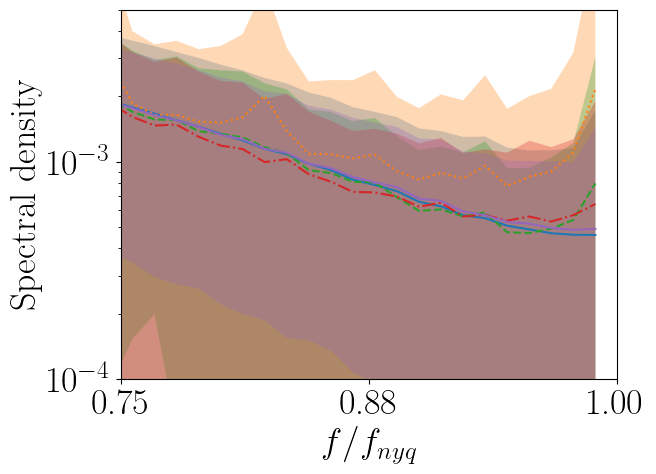}
    \raisebox{1.0cm}{
    \includegraphics[width=0.17\linewidth]{gfx/gan/ffhq_64/compare_discs_legend}}
    \caption{\textbf{Reduced Spectrum for SNGAN with Discriminators on Different Input Domains} on Cats128. We plot the mean and standard
deviation of the reduced spectrum above $0.75f_{nyq}$.}
  \label{fig:sngan_discriminators}
    \vspace{-0.3cm}
\end{figure}
Consistent with our analysis on PGAN~\cite{Karras2018ICLR},~\figref{fig:sngan_discriminators} shows that the additional spectral discriminator is the most effective to reduce the spectral discrepancies. 
However, interestingly, the corresponding classification accuracies on the reduced spectra in~\tabref{tab:discriminator_sngan} are similar for wavelets and the additional spectral discriminator. This suggests that training the spectral classifier on the fitted decay parameters, as proposed in~\cite{Dzanic2020NIPS}, can occasionally produce overly optimistic results with an accuracy closer to chance, \ie, $50\%$.
However, for all results in the main paper the classification accuracy is consistent with the qualitative alignment of the reduced spectra, see~\figref{fig:pgan_discriminators},~\figref{fig:stylegan_spectra}, and Fig. 7 in the main paper.
To adhere to the same metric as~\cite{Dzanic2020NIPS,Chandrasegaran2021CVPR}, we hence decide to report the classification accuracy on the fitted decay parameters.\\
While the FID in~\tabref{tab:discriminator_sngan} is mixed for the different approaches, all values are lower than the corresponding values for PGAN in Table 3 of the main paper. This is expected, as our version of PGAN with the reduced channels has significantly fewer parameters than the original SNGAN.

\subsection{StyleGAN2}\label{sec:stylgegan}
\begin{figure}[t!]
    \centering
    \begin{subfigure}{0.27\linewidth}
    	\includegraphics[width=\linewidth]{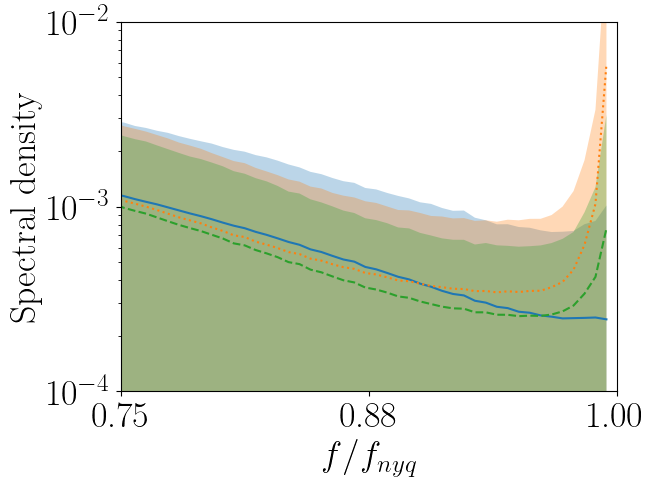}
    	\caption{Cats256}
    \end{subfigure}
    \begin{subfigure}{0.27\linewidth}
    	\includegraphics[width=\linewidth]{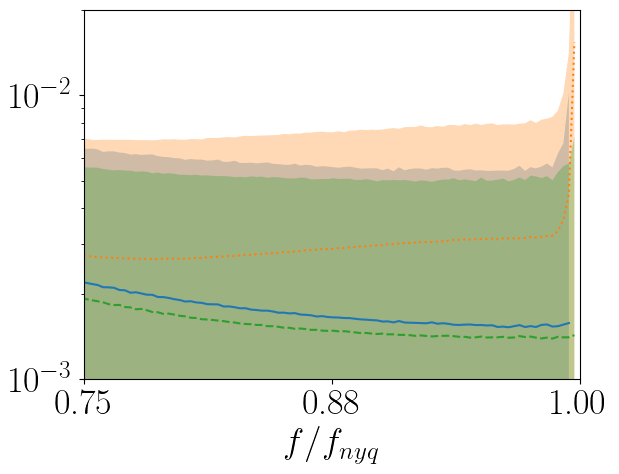}
    	\caption{AFHQ Dog}
    \end{subfigure}
    \begin{subfigure}{0.27\linewidth}
    	\includegraphics[width=\linewidth]{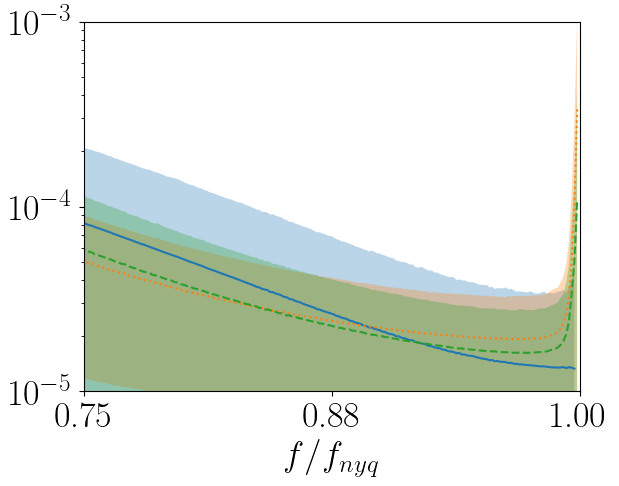}
    	\caption{FFHQ}
    \end{subfigure}
    \begin{subfigure}[b]{0.14\linewidth}
    	\includegraphics[width=\linewidth]{gfx/gan/toyset_spectrum_64/spectrum_legend}
    \end{subfigure}
    \caption{\textbf{Reduced Spectrum for StyleGAN2.} We plot the mean and standard
deviation of the reduced spectrum above $0.75f_{nyq}$.
While the spectral discriminator removes the high-frequency peak on AFHQ Dog, the spectra for the remaining datasets retain an elevated amount of high frequencies.}
  \label{fig:stylegan_spectra}
    \vspace{-0.3cm}
\end{figure}
For finetuning StyleGAN2~\cite{Karras2020CVPR} with the spectral discriminator~\cite{Jung2021AAAI}, ~\figref{fig:stylegan_spectra} shows spectrum plots corresponding to Table 5 in the main paper. 
While the spectral discriminator largely improves the spectral statistics on AFHQ Dog, it cannot fully resolve the peak at the highest frequencies for the remaining datasets. This also reflects in the high accuracy of the spectral classifier on these datasets in Table 5 of the main paper.
\begin{figure}[t!]
    \centering
    \begin{subfigure}{0.7\linewidth}
 \setlength\tabcolsep{0.1em}
    \begin{tabular}{ccc}
    	\includegraphics[height=0.3\linewidth]{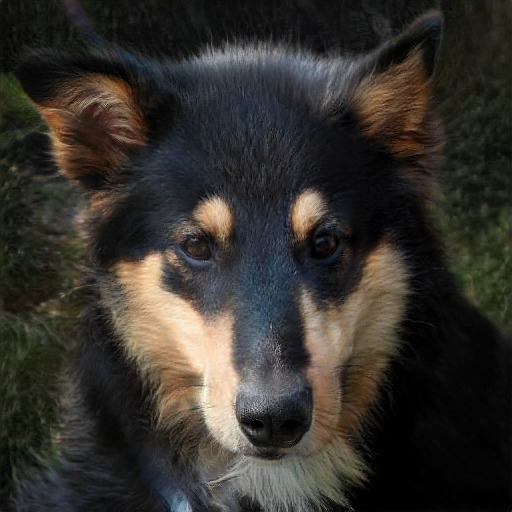} & 
    	\includegraphics[height=0.3\linewidth]{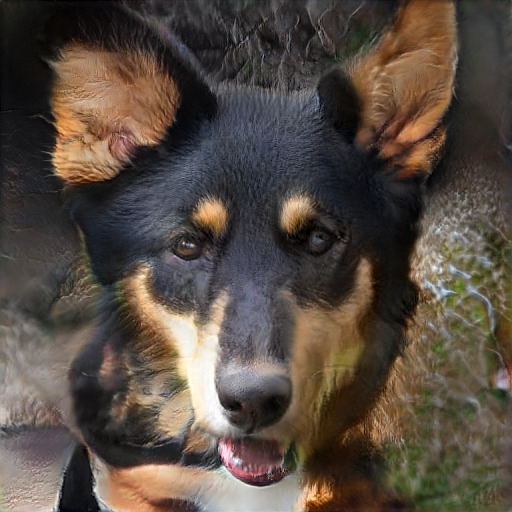} &
    	\includegraphics[width=0.4\linewidth]{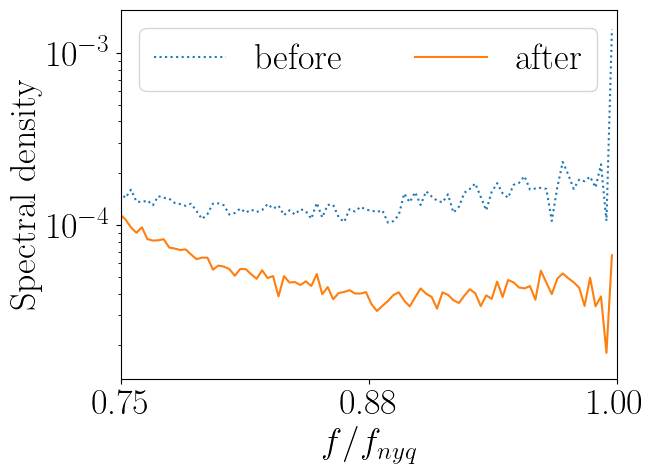}\\
    	\includegraphics[height=0.3\linewidth]{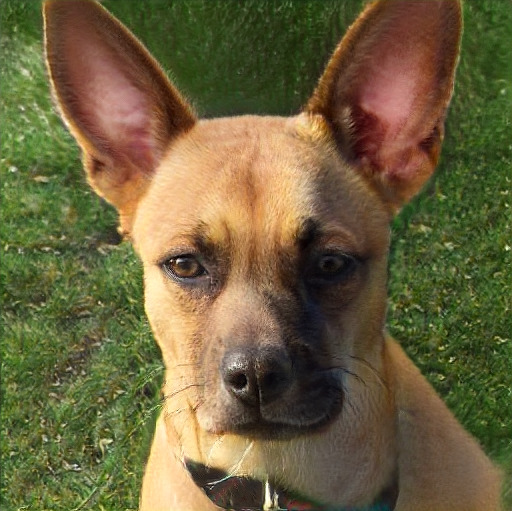} & 
    	\includegraphics[height=0.3\linewidth]{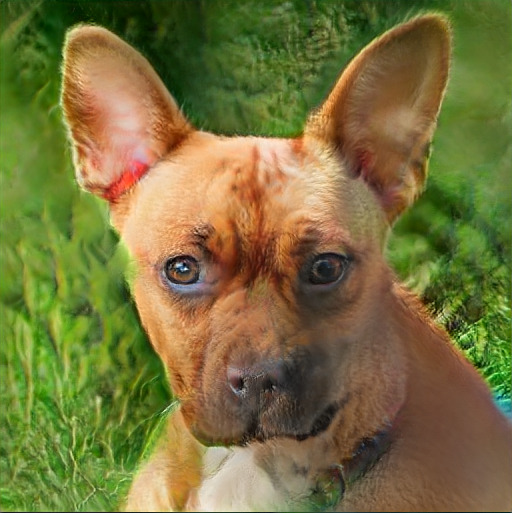} &
    	\includegraphics[width=0.4\linewidth]{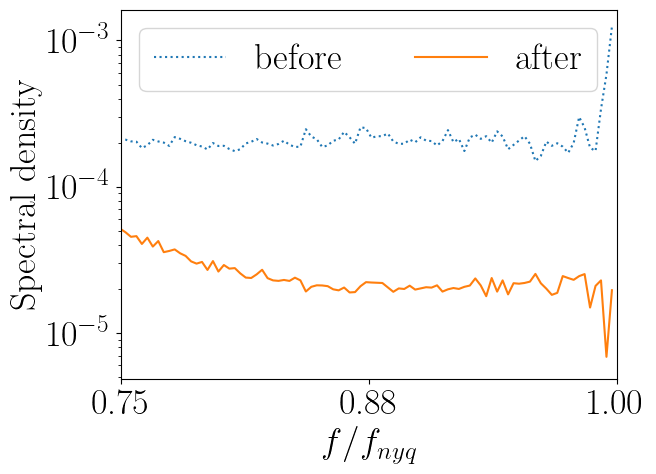}\\
    	Before & After & Reduced Spectrum
    \end{tabular}    
    	\caption{AFHQ Dog}
    \end{subfigure}
    \begin{subfigure}{0.7\linewidth}
     \setlength\tabcolsep{0.1em}
    \begin{tabular}{ccc}
    	\includegraphics[height=0.3\linewidth]{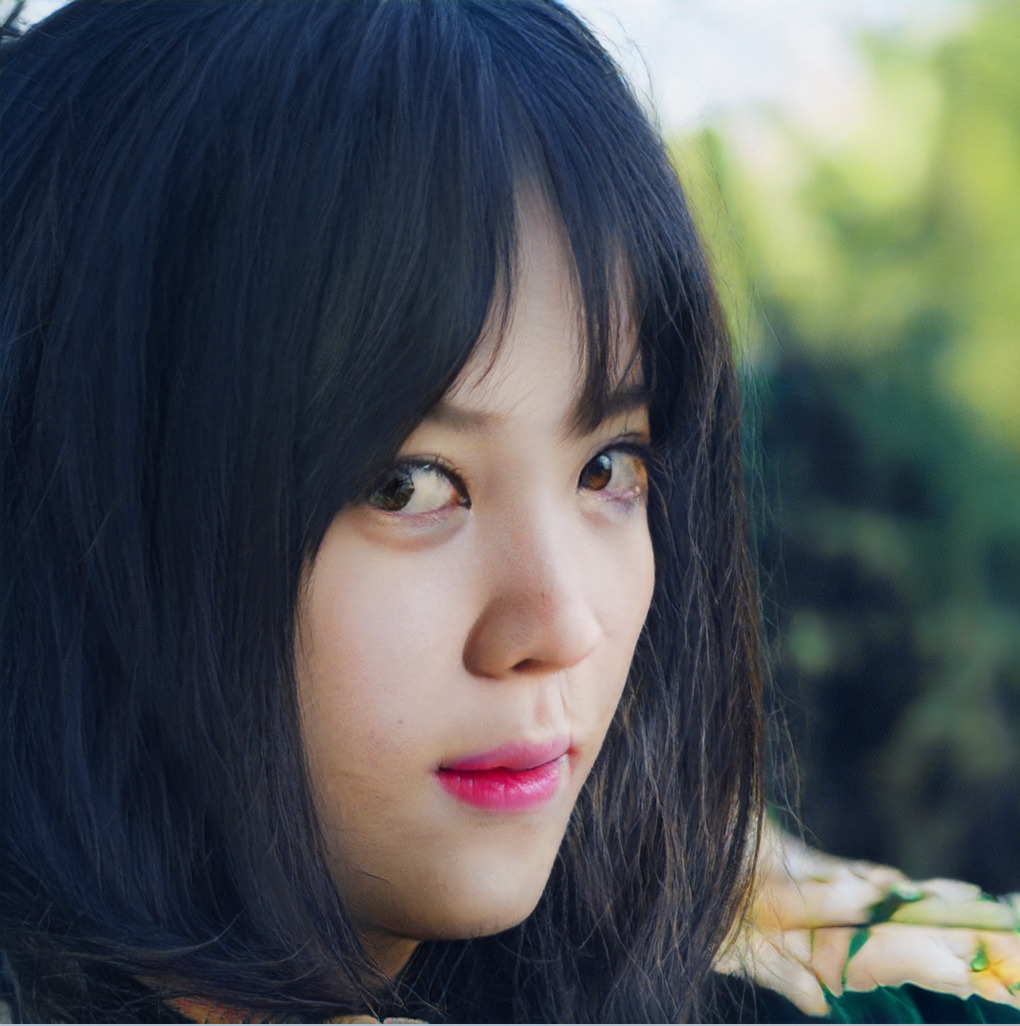} & 
    	\includegraphics[height=0.3\linewidth]{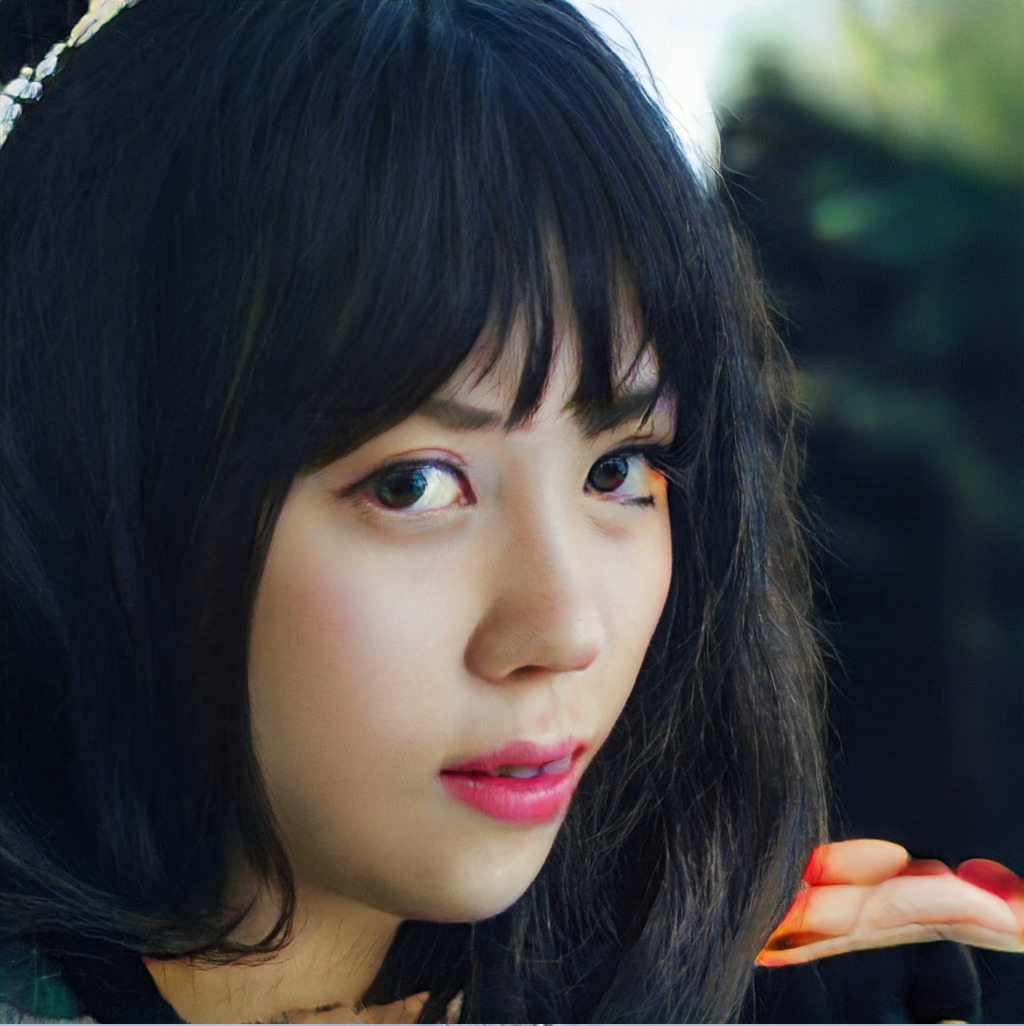} &
    	\includegraphics[width=0.4\linewidth]{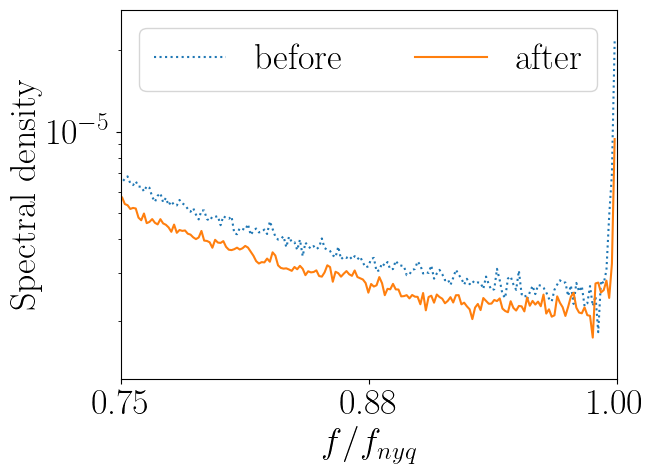}\\
    	\includegraphics[height=0.3\linewidth]{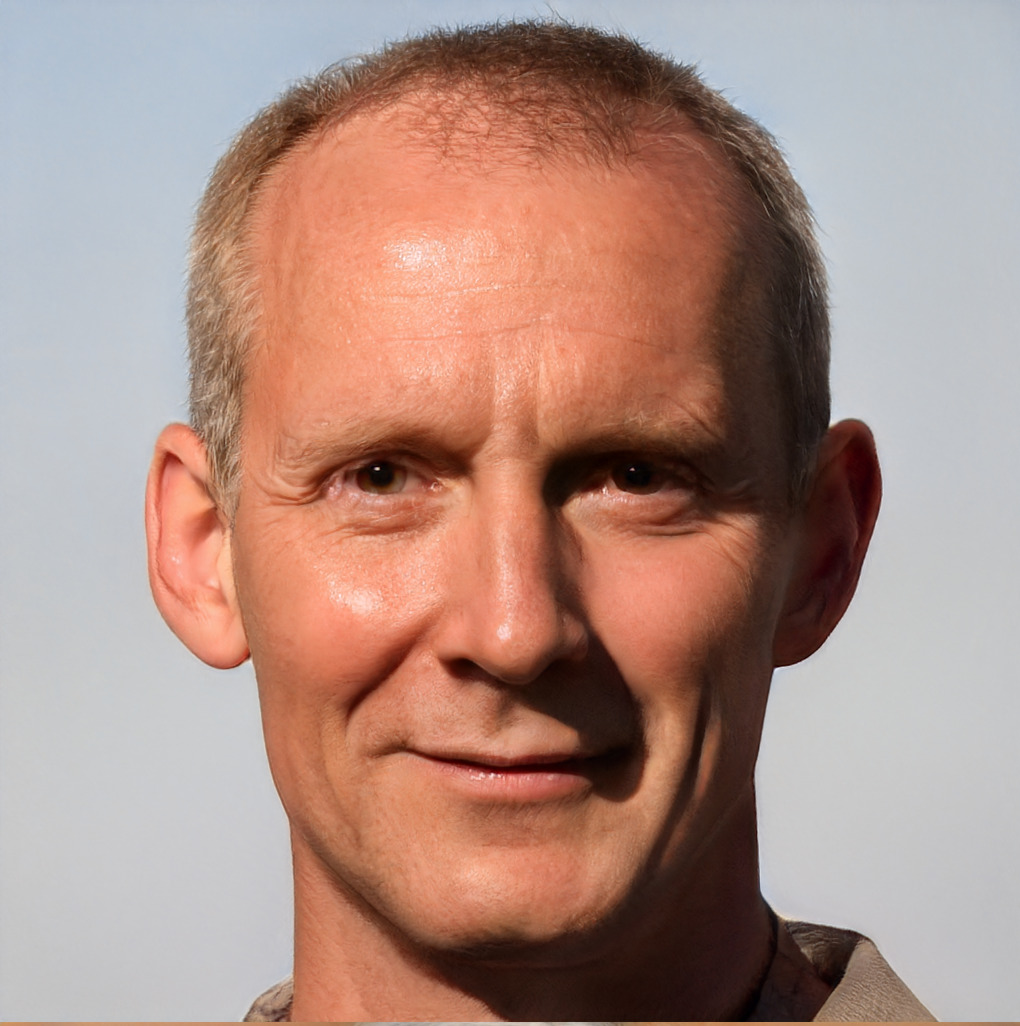} & 
    	\includegraphics[height=0.3\linewidth]{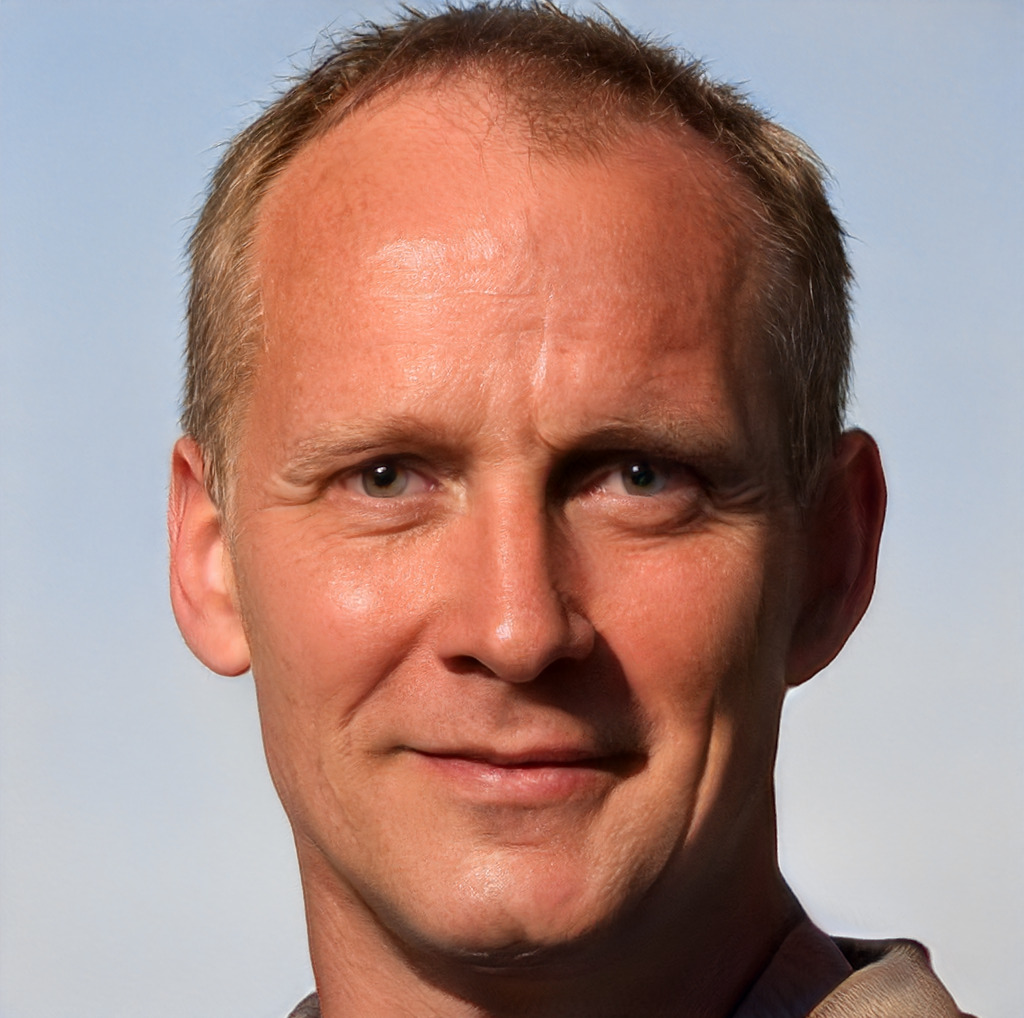} &
    	\includegraphics[width=0.4\linewidth]{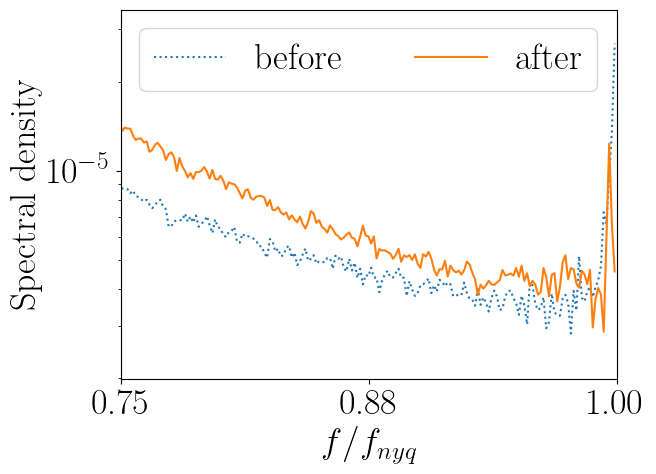}\\
    	Before & After & Reduced Spectrum
    \end{tabular}
    	\caption{FFHQ}
    \end{subfigure}    
    \caption{\textbf{Samples from StyleGAN2} and their spectra before and after finetuning with an additional discriminator on the reduced spectrum. (a) The peak at the highest frequencies is removed but image quality degrades. (b) The image fidelity remains high but the high-frequency artifacts also persist.}
  \label{fig:finetune_qualitative}
    \vspace{-0.3cm}
\end{figure}
~\figref{fig:finetune_qualitative} shows qualitative results before and after finetuning on AFHQ dog and FFHQ. For AFHQ dog, finetuning prevailingly changes the background to correct the spectral statistics but qualitatively this reduces the image fidelity. This supports the results in Table 5 in the main paper, where the accuracy of the spectral classifier is reduced but FID becomes worse.
The finetuned images on FFHQ have no background artifacts and have a similar FID as the original images. However, the spectral statistics improve only slightly, see~\figref{fig:stylegan_spectra}.
This again suggests that the reduced spectrum might not contain enough information to correct both the spectral statistics and image fidelity.

%% file: supplementary/sec_datasets.tex
\section{Datasets}\label{sec:datasets}
\boldparagraphnospace{Toyset}
We will include the scripts for generating our Toyset in our code release upon acceptance.
\boldparagraphnospace{Licenses}
The datasets used in this paper, CelebA~\cite{Liu2015ICCVa}, FFHQ~\cite{Karras2019CVPR} (Creative Commons BY-NC-SA 4.0), LSUN Cats~\cite{Yu2015ARXIV}, and AFHQ~\cite{Choi2020CVPR} (Creative Commons BY-NC 4.0) are available for non-commercial research purposes and are therefore suitable for our work. 